\pdfoutput=1

\documentclass[11pt]{article}

\usepackage{emnlp2021}

\usepackage{times}
\usepackage{latexsym}

\usepackage[T1]{fontenc}

\usepackage[utf8]{inputenc}

\usepackage{microtype}
\usepackage{amsmath}
\usepackage{multirow}
\usepackage{graphicx}
\usepackage{hhline}
\usepackage{verbatim}
\usepackage[normalem]{ulem}
\useunder{\uline}{\ul}{}
\usepackage{caption}
\usepackage{subcaption}
\usepackage{float}
\newcommand{\scqa}{$(\mathcal{SC})^2\mathcal{QA}$}
\title{Generating Self-Contained and Summary-Centric Question Answer Pairs via Differentiable Reward Imitation Learning}

\author{Li Zhou \and Kevin Small \and Yong Zhang \and Sandeep Atluri \\
Amazon Alexa\\
\{lizhouml,smakevin,yonzhn,satluri\}@amazon.com}

\begin{document}
\maketitle
\begin{abstract}
Motivated by suggested question generation in conversational news recommendation systems, we propose a model for generating question-answer pairs (QA pairs) with self-contained, summary-centric questions and length-constrained, article-summarizing answers. We begin by collecting a new dataset of news articles with questions as titles and pairing them with summaries of varying length. This dataset is used to learn a QA pair generation model producing summaries as answers that balance brevity with sufficiency jointly with their corresponding questions. We then reinforce the QA pair generation process with a differentiable reward function to mitigate exposure bias, a common problem in natural language generation. Both automatic metrics and human evaluation demonstrate these QA pairs successfully capture the central gists of the articles and achieve high answer accuracy.\footnote{Code and Data will be made available at \url{https://github.com/amazon-research/SC2QA-DRIL}}

\end{abstract}

\section{Introduction}

Automatic generation of question-answer pairs (QA pairs) is a widely studied problem, primarily used to improve the performance of question answering systems via data augmentation~\citep{alberti-etal-2019-synthetic,shakeri-etal-2020-end}. However, question generation has also recently garnered interest in the context of conversational agents, where suggested questions (SQs) (i.e., {\em You can also ask...}) have emerged as a promising approach to drive multi-turn dialogues by educating customers about the agent capabilities and guiding users along dialogue trajectories with more engaging content~\citep{yin2020question,NouriSFW20}. 

As an example, consider a news chatbot engaged in a dialogue regarding COVID-19 vaccine developments producing the SQ \{{\bf Q}: \textit{How effective is the Pfizer-BioNTech vaccine?}\} paired with the answer \{{\bf A}: \textit{Pfizer/BioNTech vaccine is around $91\%$ effective at preventing COVID-19, according to updated trial data. Experts fear new variants of COVID-19 from South Africa and Brazil may be resistant to existing vaccines and treatment.}\} Firstly, SQs of this form mitigates the user burden regarding the necessity of both deep subject knowledge to ask good questions and awareness of the agent question answering capabilities to expect good answers. Secondly, the agent can look-ahead when selecting SQs to bias toward confidently correct answers and content expected to lead to further follow-up questions and general system engagement.

Targeting the SQ problem in news chatbot scenarios (e.g., \citep{laban-etal-2020-whats}), this work examines QA pair generation corresponding to a news article summary paired with a self-contained question. Table \ref{tb:sqexample} shows an example of the task. SQs based on these summary-centric QA pairs act as implicit article recommendations, complementing SQs focusing on passage-level extracted answers or factoid information.
%
QA pairs generated for this purpose must satisfy several criteria including: (1) questions are self-contained (i.e., users need not read the corresponding articles nor require significant additional domain knowledge to unambiguously understand the questions~\citep{yin2020question}), (2) questions are summary-centric (questions capture the gists of the corresponding articles), (3) answers correctly answer the questions, and (4) answers are brief but sufficient such that users can confidently trust the results. Additionally, to support different settings (e.g., screened device, mobile device, voice-only), we explore QA pair generation for varying application-specific answer length requirements. 

\begin{table}[h]
\centering
\resizebox{0.9\linewidth}{!}{
\begin{tabular}{|p{\linewidth}|}
\hline
\textbf{Article}: \textit{President Biden’s infrastructure plan calls for an unprecedented boost in federal aid to the nation’s passenger rail system, seeking to address Amtrak’s repair backlog, extend service to more cities and modernize the network in the Northeast Corridor.
The American Jobs Plan announced Wednesday calls for \$80 billion for rail -- money that could be crucial in taking passenger service to cities such as Las Vegas and Nashville, and expand operations across large metropolitan areas such as Atlanta and Houston. "President Biden's infrastructure plan is what this nation has been waiting for," Amtrak chief executive William J. Flynn said, while echoing Biden’s push to rebuild and improve...} \\ \hline
\textbf{Suggested Question}: \textit{What does President Biden's infrastructure plan mean for Amtrak?} \\
\textbf{Short Answer}: \textit{The federal funding would help Amtrak accomplish long-needed upgrades to tracks, tunnels and bridges in the Northeast.} \\
\textbf{Long Answer}: \textit{The American Jobs Plan announced Wednesday calls for \$80 billion for rail. The federal funding would help Amtrak accomplish long-needed upgrades to tracks, tunnels and bridges in the Northeast, the nations busiest rail corridor. Amtrak has a \$45.2 billion backlog of projects that it says are needed to bring its assets to a state of good repair.} \\ \hline
\end{tabular}
}
\caption{The suggested QA pair generation task. Given an article, we generate a self-contained and summary-centric question and a length-constrained answer. The question captures the gist of the article and can be understood without reading the corresponding article.}
\label{tb:sqexample}
\vspace{-0.2cm}
\end{table}

To satisfy these requirements, we first collect a corpus of suitable QA pairs, accomplished by curating a set of news articles with well-formed questions as their titles and for which we can confidently generate variable length summaries as answers. Observing that the {\em summary generation} $\rightarrow$ {\em question generation} pipeline suffers from exposure bias~\citep{ranzato2015sequence}, we propose a novel differential reward imitation learning (DRIL) training method that samples summary answers and reconstructs questions exclusively based on the hidden states of the answer decoder. Generated summaries are capable of directly reconstructing the questions, making them more likely the answers to the questions, and generate questions more closely related to the gists of the articles. We empirically validate the model with automated and human evaluations.

In this paper, we study QA pair generation corresponding to variable length article-summarizing answers paired with self-contained and summary-centric questions. Our contributions include: (1) We collect a new QA dataset targeted for producing SQs in a news chatbot. (2) We propose a QA pair generation model where both questions and answers are well-formed, questions capture the central gists of articles, and answers are succinct while containing sufficient supporting context. (3) We propose a novel differentiable reward imitation learning (DRIL) method which shows better performance over maximum likelihood estimation (MLE) and reinforcement learning (RL) for QA pair generation. (4) We perform extensive empirical evaluations to quantify DRIL-based QA pair generation improvements.

\section{Related Works}
\textbf{Question-only Generation (QG)}. Both heristic-based~\citep{heilman-smith-2010-good} and neural models~\citep{du-etal-2017-learning,zhou2017neural,sun-etal-2018-answer} have been applied to QG. Usually, neural QG models are given contexts containing answers beforehand, contrasting with our goal of jointly generating QA pairs. \citet{tuan2020capturing,song-etal-2018-leveraging,zhao-etal-2018-paragraph} proposed to generate questions from long text and wider contexts, which is related to our method for QG using summaries. However, these wider contexts are only used to improve QG for the specified answer spans and do not attempt to capture the central gists of articles. 

\noindent \textbf{Question and Answer Generation (QG+AG)}. QG+AG generates QA pairs jointly~\citep{liu2020asking,alberti-etal-2019-synthetic,du-cardie-2018-harvesting,du-cardie-2017-identifying,subramanian-etal-2018-neural,wang2019multi,krishna-iyyer-2019-generating}, frequently with two independent steps: identify question-worthy answer spans followed by generating answer-aware questions. Recent works train neural models to generate QA pairs~\citep{shakeri-etal-2020-end,lee-etal-2020-generating} using QA datasets such as SQuAD~\citep{rajpurkar-etal-2016-squad} and Natural Questions~\citep{kwiatkowski-etal-2019-natural} modulo the goal of generating self-contained questions paired with succinct but sufficient article-summarizing answers.

\noindent \textbf{Applications of QG and QG+AG}. QG and QG+AG have been used for applications including data augmentation for QA systems~\citep{alberti-etal-2019-synthetic,shakeri-etal-2020-end}, information seeking in chatbots~\citep{qi-etal-2020-stay,laban-etal-2020-whats}, document understanding~\citep{krishna-iyyer-2019-generating}, educational practice and assessment~\citep{le2014automatic}, and online shopping~\citep{yu-etal-2020-review}.

\noindent \textbf{Training Mechanism for Sequence Prediction}. Sequence prediction models are commonly trained with MLE. However, MLE can lead to degeneration~\citep{holtzman2019curious} caused by exposure bias~\citep{ranzato2015sequence}. Many algorithms~\citep{yu2017seqgan,NIPS2016_16026d60,song2020improving,welleck2019neural} have been proposed to mitigate exposure bias. Our DRIL method not only mitigates exposure bias, but also optimizes for a differentiable reward function that is aligned with the end goal. Please refer to Section \ref{sec:il} for comparison between DRIL and existing algorithms.

\section{\scqa: A Self-Contained and Summary-Centric QA Dataset}
\label{sec:dataset}
While multiple QA datasets exist to train a QG or AG model, none specifically fit the goal of this paper. QA pairs in SQuAD~\citep{rajpurkar-etal-2018-know}, NewsQA~\citep{trischler-etal-2017-newsqa}, and Natural Questions (NQ)~\citep{kwiatkowski-etal-2019-natural} are not designed to capture the article gists, and a significant number of questions in SQuAD and NewsQA are not self-contained.

A key observation enabling this work is that many news articles have questions as their titles (e.g.\ \textit{How has the Biden administration helped student loan borrowers?}) that can be used to train a SQ generation model since these questions usually correspond to the central gists of the news articles and are designed to be understood without reading the articles. However, two challenges remain: (1) clickbait titles need to be filtered, and 2) these questions are not paired with summary-centric answers. Therefore, we developed the following data collection procedure to produce \scqa, our self-contained summary-centric QA dataset.

\subsection{Question-Article Pairs Collection}
Starting with a curated URL list of news websites, we collected all articles between September $2020$ to March $2021$ with a title that starts with a pre-defined list of words (e.g., \textit{Where}, \textit{What},  \textit{How}) and ends with a question mark.
We then define a set of rules to filter out ill-formed and clickbait titles (details in Appendix A). Finally, we remove any questions that appear in the articles to ensure we don't learn to copy the questions when present. In total, we collected $39{,}460$ such question-article pairs.

\subsection{\{Question, Article, Summary, Length Constraint\} 4-Tuples Collection}
Given collected question-article pairs, we must pair them with suitable answers to produce QA pairs. From a  preliminary study, we observed that $\sim70\%$ of title questions can be answered by summaries of the corresponding articles. As a result, we set out to augment the question-article dataset with generated summaries as pseudo ground truth answers using following three-step procedure: \\
\noindent \textbf{Step 1} (Define desired answer lengths): One of our goals is to generate well-formed answers that are succinct while containing sufficient supporting context. Therefore, we generate summaries with varying brevity. Analyzing the average number of tokens for the first $1$, $2$ and $3$ sentences of the  CNN/DailyMail summaries~\citep{NIPS2015_afdec700}, we define three buckets of varying answer lengths: $(0, 30], (30, 50]$ and $(50, 72]$ BPE tokens. \\
\noindent \textbf{Step 2} (Generate summary): For each article and desired length bucket, we use three SoTA summarization models (PEGASUS~\citep{zhang2020pegasus}, BART~\citep{lewis-etal-2020-bart}, and CTRLSum~\citep{he2020ctrlsum}) fine-tuned on CNN/DailyMail to generate three candidate summaries
-- enforcing summary length via control of \texttt{EOS} token generation. Unfinished sentences are removed and the length bucket is reassigned if needed. \\
\noindent \textbf{Step 3} (Filter-out incorrect summary answers): Not all questions can be answered by the generated summaries since: (1) even the ground truth summary may not be a correct answer to the question and (2) summaries generated by SoTA models may not be good. To identify if a candidate summary answers the question, we train a QA pair classifier using the $4$ million question-snippet pairs MSMARCO dataset~\citep{bajaj2016ms}. For each article and length bucket, we select the candidate summary that has the highest score predicted by the trained classifier. 
In total, we produce $53{,}746$ 4-Tuples of \{Question, Article, Summary, Length Constraint\}. For additional details and dataset statistics, please refer to Appendix A.

\section{Models for QA Pair Generation}
In this section, we propose a family of QA pair generation models that are trained on the data collected in Section \ref{sec:dataset}. Let $D$ denote a document (news article), $S$ denote a summary, $Q$ denote a question, $L$ denote a length bucket indicator (\texttt{LB0}, \texttt{LB1} or \texttt{LB2}), and \texttt{<s>} and \texttt{</s>} denote the special \texttt{BOS} and \texttt{SEP} tokens  respectively.

\subsection{Base D$\rightarrow$S$\rightarrow$Q Model (D-S)}
\label{sec:basemodel}
\begin{figure}[htp!]
\centering
\includegraphics[width=0.8\linewidth]{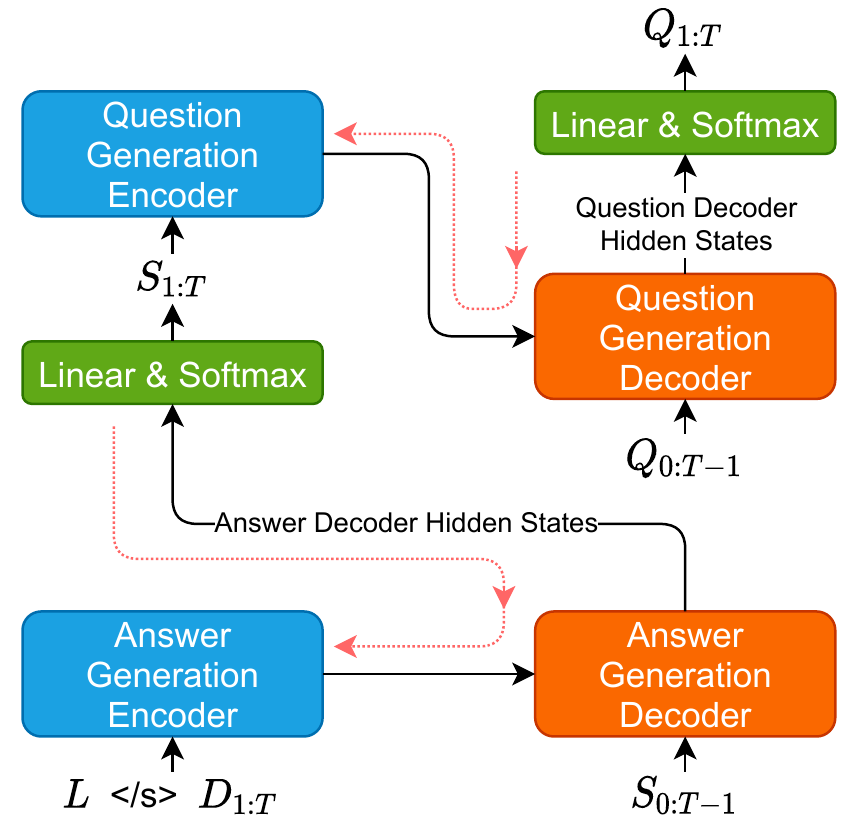}
\caption{Training of answer generation (AG) and question generation (QG) of the \textbf{D-S} model. L, D, S, Q denotes the length bucket indicator, document, summary, and question, respectively. Red dash arrows denote gradient flow.}
\label{fig:basemodel}
\vspace{-0.2cm}
\end{figure}

Our base model is shown in Figure \ref{fig:basemodel}, consisting of two transformer-based encoder-decoder models~\citep{NIPS2017_3f5ee243} where one performs answer generation (AG) and the other question generation (QG). During training, the AG model encodes a concatenation of the length bucket indicator and the document, and decodes a length-constrained summary: 
\begin{align*}
f_{\theta^a_{\text{enc}}}:& \ L + D \rightarrow c^a_{\text{enc}} \\
f_{\theta^a_{\text{dec}}}:& \ S_{0:T-1}, c^a_{\text{enc}} \rightarrow S_{1:T} 
\end{align*}
where $\theta^a_{\text{enc}}$ and $\theta^a_{\text{dec}}$ are the encoder and decoder parameters, $c^a_{\text{enc}}$ is the sequence of hidden states at the last encoder layer, $S_{1:T}$ is the ground truth summary, and $S_{0:T-1}$ is the decoder input ($S_{1:T}$ offset by one timestamp and prepended by a \texttt{BOS} token). 
The AG model is trained using MLE: 
\begin{align*}
    \mathcal{L}(\theta^a_{\text{enc}}, \theta^a_{\text{dec}}) = - \sum_{n=1}^N\log p(S^{(n)} | L^{(n)}+D^{(n)})
\end{align*}
where $(n)$ represents the $n$-th training instance. 
QG is also trained via MLE, mapping an input summary to a question: 
\begin{align*}
f_{\theta^q_{\text{enc}}}:& \ S \rightarrow c^q_{\text{enc}} \\
f_{\theta^q_{\text{dec}}}:& \ Q_{0:T-1}, c^q_{\text{enc}} \rightarrow Q_{1:T} \\
\mathcal{L}(\theta^q_{\text{enc}}, \theta^q_{\text{dec}}) &= - \sum_{n=1}^N\log p(Q^{(n)} | S^{(n)})
\end{align*}

During inference, when decoding summary answers, we again control the generation of \texttt{EOS} to fall into the range specified by the desired length bucket. We remove any unfinished sentences at the end unless after the truncation the answer is shorter than the minimum length of the length bucket.

We use a pre-trained BART model~\citep{lewis-etal-2020-bart} to initialize $\theta^a_{\text{enc}}$, $\theta^a_{\text{dec}}$, $\theta^q_{\text{enc}}$ and $\theta^q_{\text{dec}}$.
We name this base model D-S since the AG model takes the document (D) as input and the QG model takes the summary (S) as input. In Section \ref{sec:variants} we will describe multiple variants of this model.

\subsection{Optimizing Answer Generation by Differentiable Rewards}
\label{sec:il}
When using MLE to train the base model, the decoder input at timestep $t$ is the ground truth token at timestep $t-1$, sometimes called teacher-forcing~\citep{williams1989learning} and known to suffer from exposure bias~\citep{ranzato2015sequence} due to the mismatch between training and inference. That is, during inference the decoder input is the predicted token instead of the ground truth token of the last timestep, causing errors from each timestamp to accumulate during generation. It has been shown that neural text generation models trained with MLE lead to generic and repetitive outputs~\citep{welleck2019neural,holtzman2019curious}. Additionally, we usually want to optimize generation metrics (e.g., ROUGE) and human feedback directly instead of optimizing training data likelihood. To mitigate these concerns, we can sample decoder output during \emph{training} and calculate the loss of the sampled output. Several works use RL to achieve this for text generation~\citep{stiennon2020learning,ziegler2019fine,yu2017seqgan} and directly optimize for preferred metrics. However, RL is not sample efficient and difficult to tune in text generation tasks due to sparse rewards. For example, \citet{hosking-riedel-2019-evaluating} have shown that applying RL to QG do not improve human evaluation metrics.

\begin{figure}[t]
\hspace{-0.4cm}
\centering
\includegraphics[width=0.8\linewidth]{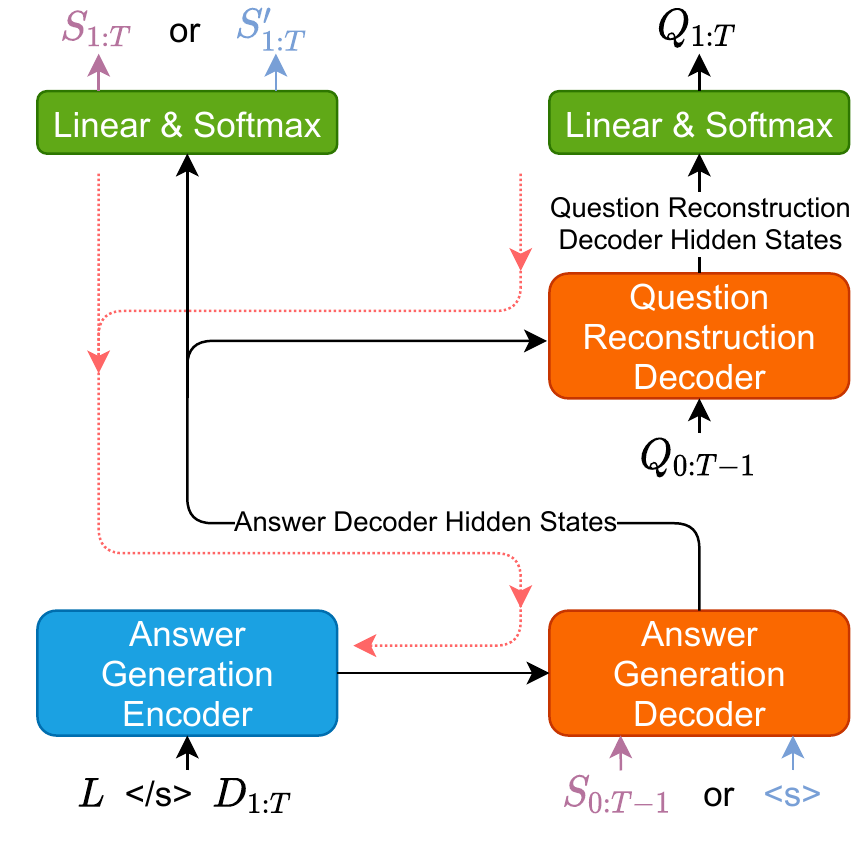}
\caption{Training of answer generation (AG) of the \textbf{D-S-DRIL} model. The input to the AG decoder is either $S_{0:T-1}$ or \texttt{<s>}. When the input is $S_{0:T-1}$, the AG decoder uses teacher-forcing to predict $S_{1:T}$, and the gradients back-propagate from $S_{1:T}$ to the AG decoder and AG encoder (the red dash arrow on the middle left), which is similar to the AG of the D-S model. However, when the input is \texttt{<s>}, the AG decoder samples a summary $S'_{1:T}$, and the answer decoder hidden states are used to reconstruct the question $Q_{1:T}$. The gradients back-propagate from $Q_{1:T}$ to the AG decoder and AG encoder (the red dash arrow on the top right). This reinforces the model to generate summaries that can reconstruct the questions.}
\label{fig:dril}
\vspace{-0.2cm}
\end{figure}

Meanwhile, we observe that when generating a summary as the answer of a QA pair, we want to generate a summary that can better reconstruct the ground truth question without the article since: (1) a summary that can reconstruct a question is more likely to be able to answer that question and (2) a summary that better reconstructs the ground truth question leads to a generated question that is closer to the gist of the article. Moreover, the AG model is conditioned on the length bucket to control the levels of brevity, meaning that when the maximum allowed answer length is short, the question reconstruction will enforces the AG model to generate \emph{succinct} but \emph{informative} answers with respect to the question given the selected brevity level.
We validate these assumptions in Section~\ref{sec:experiments}.

We now propose the differentiable reward imitation learning (DRIL) method for training the AG model as shown in Figure \ref{fig:dril}.
 During training, the AG model performs \textit{vanilla} sampling to generate a summary: 
\begin{align*}
f_{\theta^a_{\text{enc}}}:& \ L + D \rightarrow c^a_{\text{enc}} \\
f_{\theta^a_{\text{dec}}}:& \ \text{\texttt{BOS}}, c^a_{\text{enc}} \rightarrow  c^a_{\text{dec}}, S'
\end{align*}
where $c^a_{\text{dec}}$ is the sequence of hidden states at the last layer of the decoder, and $S'$ is the sampled summary. This differs from teacher-forcing since summaries are sampled in training. We then use another transformer-based decoder to reconstruct the question:
\begin{align*}
f_{\theta^r_{\text{dec}}}: Q_{0:T-1}, c^a_{\text{dec}} \rightarrow Q_{1:T}
\end{align*}
noting that this decoder only depends on the hidden states of the AG decoder (not $L + D$). This forces the model to reconstruct the question only from the summary.
The gradient can back-propagate from the question to the hidden states of the AG decoder $c^a_{\text{dec}}$ and AG encoder $c^a_{\text{enc}}$ such that the question reconstruction loss will guide AG. To ensure generated summary fluency, we also add the MLE loss from the base model. Overall, the AG model's loss function is given by:
\begin{align*}
    \mathcal{L}(\theta^a_{\text{enc}}, \theta^a_{\text{dec}}, &\theta^r_{\text{dec}}) = \\
     -\sum_{n=1}^N &\lambda \log p(Q^{(n)} | S'^{(n)}, L^{(n)} + D^{(n)}) \\
     & + (1-\lambda) \log p(S^{(n)} | L^{(n)}+D^{(n)})
\end{align*}
In our experiments, $\lambda=0.3$ performs the best on the validation set. 
Finally, while we apply DRIL to the training of the AG model, the QG model remains the same as the base model. We do not use the question reconstruction decoder  $\theta_{dec}^r$ as our QG model because its encoder input $c^a_{dec}$ is a uni-directional representation and hence not preferred. We call this QA pair generation model D-S-DRIL.

\textbf{Connection with RL, Unlikelihood~\citep{welleck2019neural}, SeqGAN~\citep{yu2017seqgan}, and Professor-forcing~\citep{NIPS2016_16026d60}, etc.} These methods mitigate exposure bias to some degree by calculating the loss from sampled sequences during training. Unlikelihood training penalizes the likelihood of undesired sampled sequences. SeqGAN and Professor-forcing both calculate the loss using a discriminator which learns to distinguish between the generated and ground truth sequences. They don't optimize an extrinsic reward function. \citet{caccia2019language} show that Language GANs suffer from mode collapse and do not outperform MLE on the quality and diversity evaluation. SeqGAN uses RL optmization and thus suffers from aforementioned issues. Our DRIL method, on the other hand, learns to optimize a differentiable reward function that \emph{aligns with the end goal}, and has lower gradient variance compared with RL. We empirically compare RL with DRIL in Section~\ref{sec:experiments}.

Beyond this work, DRIL can be applied to other sequence prediction problems. For example, in step-by-step instruction following such as ALFRED tasks~\citep{shridhar2020alfred}, DRIL can optimize the current step's action trajectory such that it can reconstruct the next $K$ instructions. The intuition is if the current step's action trajectory is correct, then the agent should be able to follow the ground truth actions in the next steps to fulfill the tasks. From this perspective, DRIL is similar to SQIL~\citep{reddy2019sqil}, which avoids drifting away from the demonstrations over long horizons by encouraging trajectories that return to demonstrated states when encountering unseen states. In conversational AI, ~\citet{NEURIPS2020_e9462095} proposed to fine-tune a GPT-2 model to generate system responses turn-by-turn. DRIL can optimize response generation at each turn such that the response and dialogue context can reconstruct the next $K$ turns' user and system response with a similar intuition: a correct system response will increase the likelihood of the ground truth in future turns. It avoids drifting away from demonstrations and mitigates exposure bias.

\subsection{Base Model Variants}
\label{sec:variants}
\begin{table*}[ht!]
\centering
\resizebox{0.8\linewidth}{!}{
\begin{tabular}{|c|c|c|c|c|c|c|c|c|}
\hline
         & \multicolumn{4}{c|}{Training}                                                     & \multicolumn{4}{c|}{Inference}                                                    \\ \cline{2-9} 
         & \multicolumn{2}{c|}{Answer Generation} & \multicolumn{2}{c|}{Question Generation} & \multicolumn{2}{c|}{Answer Generation} & \multicolumn{2}{c|}{Question Generation} \\ \cline{2-9} 
            & Encoder             & Decoder          & Encoder             & Decoder            & Encoder             & Decoder          & Encoder             & Decoder            \\ \hline
D-S         & L+D                 & S                & S                   & Q                  & L + D               & S'               & S'                  & Q'                 \\ \hline
D-D         & L + D               & S                & D                   & Q                  & L + D               & S'               & S'                  & Q'                 \\ \hline
D-SD        & L + D               & S                & S + D               & Q                  & L + D               & S'               & S' + D              & Q'                 \\ \hline
QD-D        & Q + L + D           & S                & D                   & Q                  & Q' + L + D          & S'               & D                   & Q'                 \\ \hline
D-S-DRIL    & L + D               & S/S'             & S                   & Q                  & L + D               & S'               & S'                  & Q'                 \\ \hline
D-S-RL      & L + D               & S/S'               & S                   & Q                  & L + D               & S'               & S'                  & Q'                 \\ \hhline{|=|=|=|=|=|=|=|=|=|}
QAGen 2S    & D + Q               & S                & D                   & Q                  & D + Q'              & S'               & D                   & Q'                 \\ \hline
\end{tabular}
}
\caption{A summary of models (D-S and its variants) we proposed for QA pair generation. Q' and S' denote the question and answer generated during inference, respectively. QAGen 2S~\citep{shakeri-etal-2020-end} is a state-of-the-art baseline. A full table that includes all the baselines in our experiments is shown at Appendix Table \ref{table:algosfull}.}
\label{table:algos}
\vspace{-0.2cm}
\end{table*}

In this section, we specify additional baseline QA pair generation models. Similar to the base D-S model, these models are based on transformer encoder-decoder architectures. The differences between these models are the encoder and decoder inputs during training and inference as summarized in Table \ref{table:algos}. Models are named by the encoder input of the AG and QG models joined with a `-'. D-D is similar to D-S except that QG takes the document (D) rather than the summary (S) as encoder input. QD-D generates question-conditioned answers, such that the AG model becomes a question-answering model. D-SD is an extension of D-S and D-D such that the encoder of the QG model takes the concatenation of S and D. D-S-DRIL optimizes the AG model of D-S using DRIL. D-S-RL optimizes the AG model of D-S using RL, and the reward function is defined as the negative question reconstruction loss calculated by the QG model of D-S. For further details, refer to Appendix B.

\section{Experiments}
\label{sec:experiments}
We conduct experiments to answer 3 research questions: (1) \textit{How good are the QA pairs generated by each algorithm?}, (2) \textit{Can DRIL outperform MLE and RL on QA pair generation?}, and (3) \textit{Is our \scqa \ dataset preferable compared with existing public QA datasets for QA pair generation?}
%
For each generated QA pair, we are interested in evaluating the following 3 questions: (1) \textit{Does the length-constrained summary answer the question?}, (2) \textit{Does the question capture the article gist?}, (3) \textit{Is the question self-contained?}
We specify automated metrics and human evaluations to quantify the answers to these research questions.
\subsection{Automated Metrics}
\textbf{ROUGE-L (R-L) and BLEU}. ROUGE-L and BLEU evaluate generated summaries/questions with respect to reference summaries/questions in the validation set. \\
\noindent \textbf{QA Pair Classifier Scores (QACS)}. We need to measure how well the generated summaries answer the \textit{generated} questions despite not having ground truth answers. Using the trained QA pair classifier from Section \ref{sec:dataset}, we propose QACS, which is the average of classifier predicted scores on the generated QA pairs. The pseudo upper and lower bounds of QACS are $0.359$ and $0.046$ based on the average classifier predicted scores of the positive and negative QA pairs in our human evaluation. 

\subsection{Human Evaluation}
We conduct human evaluation on Amazon Mechanical Turk. We designed 7 annotation tasks (ATs). Please refer to Appendix C for detailed human evaluation setup. Here we describe 4 ATs for which we are most concerned: \textbf{AT-1} shows a QA pair and asks \textit{Without referring to the answer or the article, are you able to understand the question?} (Is the question self-contained?), \textbf{AT-2} follows AT-1 and asks \textit{Does the passage in the Answer text box answers the question?}, \textbf{AT-5} shows the corresponding article and asks \textit{Does the question in the Question text box capture the gist of the Article?}. For these three tasks, annotators select either \textsc{True} or \textsc{False}. \textbf{AT-6} shows an article and a list of questions generated by different models and asks \textit{Which Question shown above best captures the gist of the Article?} 

\subsection{Baseline}
We evaluate D-S and its variants in Table \ref{table:algos}. Beyond that, we evaluate the following baselines. \textbf{QAGen 2S:} This is the state-of-the-art model for QA pairs generation for improving QA systems. We train QAGen 2S on our dataset, which is similar to QD-D except that there is no length control on the answers. \textbf{CTRLSum:} We use a pre-trained CTRLSum model to generate question-dependent summaries. Questions are generated by the QG model of QD-D. \textbf{QA Transfer:} We train a question-answering model on the NewsQA dataset to answer the generated questions. Questions are generated by the QG model of QD-D. This is to verify if a pre-trained question-answering model is sufficient to answer the questions in our dataset. \textbf{D-S-NewsQA} and \textbf{Natural Questions (D-S-NQ):} These two models are similar to D-S, except that the QG models are trained on NewsQA and NQ, respectively. This is to verify if \scqa \ is better than other existing QA datasets for QG tasks. Refer to Appendix B for implementation details.

\subsection{Data}
\textbf{Training and Validation set}.
We use the data described in Section \ref{sec:dataset} for training, taking the last $5{,}000$ out of the $53{,}746$ examples as validation set.

\noindent \textbf{Test set}.
It is desirable to evaluate models on articles that do not use questions as titles.
We sampled news articles between April $1$ to April $7$, $2021$ from the following news domains: washingtonpost.com, reuters.com, foxnews.com, cnn.com, cbsnews.com, nbcnews.com, nydailynews.com. We filtered out articles that use questions as titles, and removed all questions in the articles. In total we collect $7{,}698$ test examples.
Unlike validation set, there are no ground truth questions or answers in the test set.

\subsection{Quality of Generated Answers}
\label{sec:qltofa}
In this section we measure the quality of answers, particularly, whether they answer the corresponding questions. 
In Table \ref{tb:qa_matching}, we show the ROUGE-L score of predicted summaries on the validation set, and QACS and AT-2 accuracy on the test set, resulting in the following observation.

\begin{table*}[]
\centering
\resizebox{\linewidth}{!}{
\begin{tabular}{|l|c|c|c|c|c|c|c|c|c|}
\hline 
 \multirow{2}{*}{}           & \multicolumn{3}{c|}{length bucket 0}                                                                       & \multicolumn{3}{c|}{length bucket 1}                                                                         & \multicolumn{3}{c|}{length bucket 2}                                                                   \\ \cline{2-10} 
            & R-L                          & QACS             & AT-2 Accuracy                                             & R-L                            & QACS             & AT-2 Accuracy                                                & R-L                          & QACS             & AT-2 Accuracy                                           \\ \hline
D-S         & \multirow{3}{*}{$59.993$}       & $0.219$          & $0.623 \pm 0.052$                                     & \multirow{3}{*}{$54.110$}         & $0.255$          & $0.748 \pm 0.045$                                     & \multirow{3}{*}{$52.236$}       & $0.294$          & $0.754 \pm 0.046$                                 \\ \cline{1-1} \cline{3-4} \cline{6-7} \cline{9-10} 
D-D         &                                 & $0.176$          & $0.571 \pm 0.053$                                     &                                   & $0.224$          & $0.683 \pm 0.048$                                     &                                 & $0.268$          & $0.782 \pm 0.045$                                 \\ \cline{1-1} \cline{3-4} \cline{6-7} \cline{9-10} 
D-SD        &                                 & $0.125$          & $0.403 \pm 0.072$                                     &                                   & $0.184$          & $0.547 \pm 0.074$                                     &                                 & $0.235$          & $0.653 \pm 0.071$                                 \\ \hline
QD-D        & {\ul $62.219$} & $0.110$          & $0.412 \pm 0.072$                                     & {\ul ${55.200}$} & $0.167$          & $0.508 \pm 0.071$                                     & {\ul $53.075$} & $0.212$          & $0.574 \pm 0.072$                                 \\ \hline
D-S-DRIL    & $58.153$                        & $\mathbf{0.225}$ & {\ul{$\mathbf{0.631 \pm 0.049}$}} & $53.376$                          & $\mathbf{0.263}$ & {\ul {$\mathbf{0.771 \pm 0.042}$}} & $50.816$                        & $\mathbf{0.304}$ & {\ul $\mathbf{0.814 \pm 0.038}$} \\ \hline
D-S-RL      & $59.466$                        & $0.224$          & $0.624 \pm 0.065$                                     & $53.635$                          & $0.262$          & $0.733 \pm 0.060$                                     & $51.871$                        & $0.302$          & $0.813 \pm 0.053$                                 \\ \hline
CTRLSum     & $48.973$                        & $0.040$          & $0.112 \pm 0.046$                                     & $52.766$                          & $0.132$          & $0.438 \pm 0.073$                                     & $50.205$                        & $0.183$          & $0.530 \pm 0.075$                                 \\ \hline
QAGen 2S    & $56.881$                        & $0.112$          & $0.412 \pm 0.073$                                     & $52.912$                          & $0.171$          & $0.536 \pm 0.072$                                     & $51.741$                        & $0.218$          & $0.611 \pm 0.071$                                 \\ \hline
QA Transfer & -                               & $0.091$          & $0.521 \pm 0.071$                                     & -                                 & $0.128$          & $0.587 \pm 0.070$                                     & -                               & $0.156$          & $0.687 \pm 0.065$                                 \\ \hline
\end{tabular}
}
\caption{Evaluation of Answer Quality. {\ul Underline}, \textbf{bold}, and {\ul \textbf{bold}} represent the best results on ROUGE-L (R-L), QACS, and human evaluation, respectively. We report a $95\%$ binomial proportion confidence interval on human evaluation. D-S-DRIL generates higher quality answers than baselines in all three answer length bucket on test set.}
\label{tb:qa_matching}
\vspace{-0.2cm}
\end{table*}

\noindent
\textbf{Models that generate questions based on answers have higher QACS and AT-2 accuracy than models that generate answers based on questions.} Recall that during inference, D-S, D-D, D-S-DRIL and D-S-RL first generate summaries as answers and then generate questions based on the answers (see Table \ref{table:algos}). These algorithms perform much better than QD-D, CTRLSum, QAGen 2S and QA Transfer which first generate questions and then generate answers to these questions. For example, D-S achieves $51.2\%$, $39.6\%$, and $23.4\%$ higher AT-2 accuracy than QAGen 2S in each of the $3$ length buckets respectively.  This observation is consistent in both QACS and AT-2 accuracy. Meanwhile, QD-D achieves the best ROUGE-L scores while the QACS and AT-2 accuracy are significantly lower than D-S (e.g., AT-2 accuracy is $33.9\%$ lower than D-S in length bucket 0). All these observations show that, to ensure the generated questions and answers match with each other, we should generate questions from answers rather than the opposite direction. This is especially true on our dataset, because the ground truth answers of our dataset are summaries, which are generated without conditioning on the questions (modulo examples generated by the CTRLSum in Section \ref{sec:dataset}).

\subsection{Quality of Generated Questions}
\subsubsection{Results on \scqa \ Dataset}
\label{sec:qltofq}
In this section, we evaluate the quality of generated questions, particularly, whether the questions capture the gists of articles. From Section \ref{sec:qltofa} we already observed that only D-D, D-S, D-S-DRIL, and D-S-RL can generate high quality answers. Therefore, here we only focus on these four models (refer to Appendix C and Section \ref{sec:overallqa} for results on other models). The results are shown in Table \ref{tb:qg_gist}. 
We report ROUGE-L/BLEU score of predicted questions on the validation set. Questions are predicted from \textit{predicted} summaries instead of ground truth summaries, which is consistent with inference on the test set where we also don't have ground truth summaries. We also report AT-5 accuracy on test set and make the following observations.

\noindent \textbf{DRIL and RL reinforce AG with question reconstruction loss and thus better reconstruct ground truth questions on validation set and better capture gists of articles on test set.} Table \ref{tb:qg_gist} shows that D-S-DRIL achieves higher ROUGE-L and BLEU score than D-S across all the length buckets. Note that D-S and D-S-DRIL have the same QG model so the only difference is the AG model, showing that D-S-DRIL is able to generate better summaries that can better reconstruct the ground truth questions. This aligns with our goal of designing the question reconstruction loss. Meanwhile, we assume that in our dataset the ground truth questions capture the gists of articles, this means that, by optimizing question reconstruction loss, D-S-DRIL can generate questions that better capture the gists of articles. This is validated by the results on AT-5 accuracy. D-S-DRIL has about $6\%$ and $3\%$ higher AT-5 accuracy than D-S on length bucket 0 and 1, respectively. D-S-DRIL has lower AT-5 accuracy than D-S on length bucket 2, likely because when the maximum allowed summary length is long, there is sufficient information to reconstruct the questions even without the reconstruction loss. D-S-DRIL also shows better performance compared with D-S-RL, indicating the advantage of differentiable question reconstruction loss over the non-differentiable question reconstruction reward.

\begin{figure}[htp!]
\includegraphics[width=\linewidth]{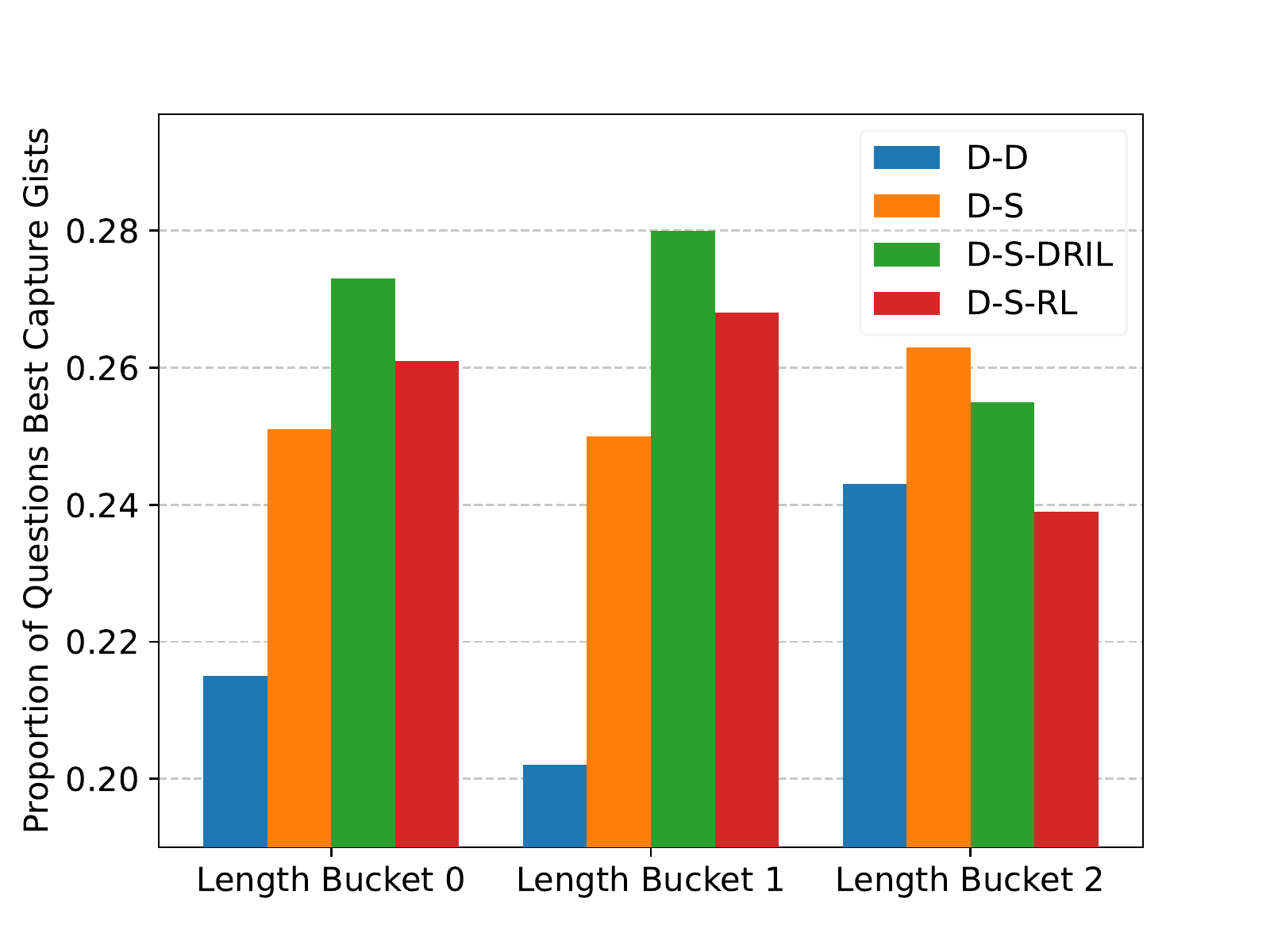}
\caption{Proportion of most preferred AT-6 questions. (\textit{Which question best captures the gist of the article?}) According to human evaluation, questions generated by D-S-DRIL best captures the gist of the article in answer length bucket 0 and 1.}
\label{fig:ht6}
\vspace{-0.2cm}
\end{figure}

AT-6 shows one article and a list of questions generated by D-D, D-S, D-S-DRIL, and D-S-RL. Annotators select the question that best captures the gist of the displayed article. Figure \ref{fig:ht6} shows the percentage of each model selected. We can see that questions generated by D-S-DRIL are preferred in length bucket 0 and 1, which is consistent with our results in Table \ref{tb:qg_gist}.

\begin{table*}[htp!]
\centering
\resizebox{\linewidth}{!}{
\begin{tabular}{|l|c|c|c|c|c|c|}
\hline
 \multirow{2}{*}{}       & \multicolumn{2}{c|}{length bucket 0} & \multicolumn{2}{c|}{length bucket 1} & \multicolumn{2}{c|}{length bucket 2} \\ \cline{2-7} 
         & R-L/BLEU      & AT-5 Accuracy    & R-L/BLEU    & AT-5 Accuracy    & R-L/BLEU    & AT-5 Accuracy    \\ \hline
D-D      & $37.274$/$9.666$      & $0.697 \pm 0.047$    & $37.605$/$10.357$     & $0.736 \pm 0.045$    & $37.643$/$10.688$     & $0.782 \pm 0.042$    \\ \hline
D-S      & $41.710$/$14.499$     & $0.768 \pm 0.043$    & $41.156$/$13.423$     & $0.782 \pm 0.042$    & $40.489$/$13.174$     & $\underline{\mathbf{0.817 \pm 0.040}}$    \\ \hline
D-S-DRIL & $\mathbf{42.764}$/$\mathbf{14.867}$     & $\underline{\mathbf{0.814 \pm 0.040}}$    & $\mathbf{41.445}$/$\mathbf{13.668}$     & $\underline{\mathbf{0.806 \pm 0.040}}$    & $\mathbf{40.678}$/$\mathbf{13.722}$     & $0.809 \pm 0.040$    \\ \hline
D-S-RL   & $42.596$/$14.756$     & $0.787 \pm 0.042$    & $40.335$/$13.100$     & $0.779 \pm 0.042$    & $40.152$/$12.906$     & $0.815 \pm 0.040$    \\ \hline
\end{tabular}
}
\caption{Evaluation of Question Quality. \textbf{Bold}, and {\ul \textbf{bold}} represents the best results on ROUGE-L(R-L)/BLEU and AT-5 accuracy, respectively. We report a $95\%$ binomial proportion confidence interval on human evaluation. D-S-DRIL generates significantly better questions in answer length bucket 0 and 1.}
\label{tb:qg_gist}
\vspace{-0.2cm}
\end{table*}

\subsubsection{\scqa \ v{.}s{.} Exiting QA Datasets}
\label{sec:ourvsqadataset}
\begin{figure}[htp!]
\includegraphics[width=\linewidth]{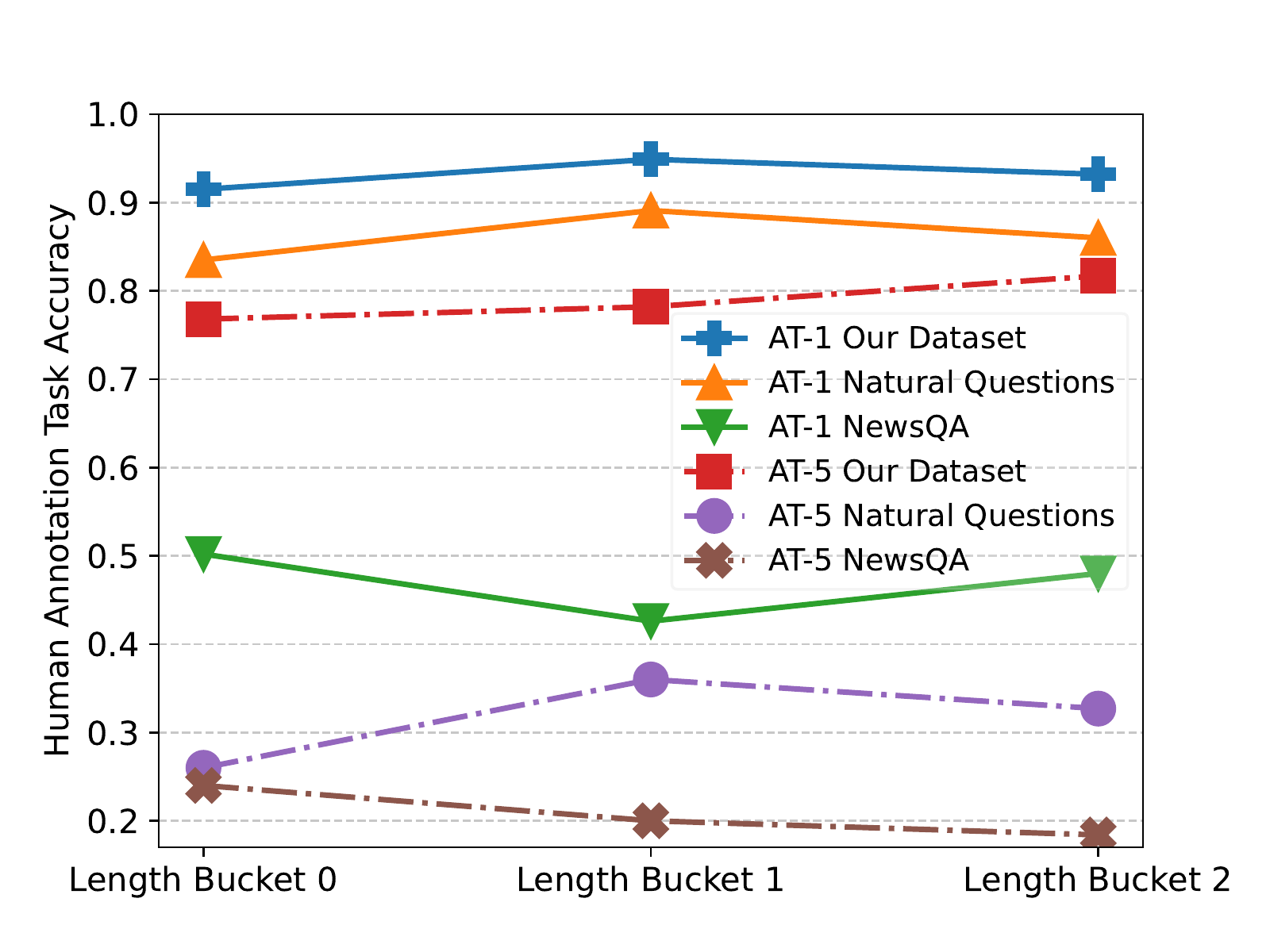}
\caption{QG Human evaluation on different datasets. Our \scqa\ dataset performs better than NewsQA and Natural Questions on both AT-1 and AT-5 human evaluation in all three answer length buckets. }
\label{fig:otherdataset}
\vspace{-0.2cm}
\end{figure}

In this section, we evaluate if \scqa \ is better than existing publicly available QA datasets for QG. We compare with D-S-NewsQA and D-S-NQ. NewsQA and NQ datasets are designed for question-answering but not QG specifically. Similar to \scqa, NewsQA is in news domain but without explicitly self-contained questions. For example, the question ``\textit{what are they going to address?}'' in the NewsQA dataset is incomprehensible without reading the article due to lack of pronoun resolution.
The human evaluation results are shown in Figure \ref{fig:otherdataset}, leading to the following observation.

\noindent \textbf{QG models trained on NewsQA and Natural Questions cannot generate self-contained questions that capture gists of articles due to the limitations of the datasets, while QG models trained on \scqa \ can.} We can see that the QG model trained on NewsQA achieves about $50\%$ lower AT-1 accuracy than the other two models, indicating that it cannot generate self-contained questions. Moreover, QG models trained on NewsQA and Natural Questions achieve $73.55\%$ and $60.03\%$ lower accuracy on AT-5 (averaged over $3$ length buckets) compared with the QG model trained on \scqa , even though all models generate questions from summaries. We observe that D-S-NewsQA tends to ask trivial questions such as the name of a person. D-S-NQ also fails to identify the focus of a summary. For example, in the summary ``\textit{Michael Jordan has two brothers and two sisters. He grew up playing basketball and baseball against his older brother.}'', D-S-NQ generates ``\textit{Who is Michael Jordan's brother playing against?}''. However, the summary focus is \textit{Michael Jordan} rather than \textit{his brother}. We discuss such cases further in the qualitative analysis section.

\subsection{Overall QA Pair Quality}
\label{sec:overallqa}
\begin{table}[]
\centering
\resizebox{\linewidth}{!}{
\begin{tabular}{|l|c|c|c|}
\hline
            & length bucket 0        & length bucket 1        & length bucket 2        \\ \hline 
D-S         & $0.566 \pm 0.036$          & $0.670 \pm 0.034$          & $0.693 \pm 0.034$          \\ \hline 
D-D         & $0.521 \pm 0.037$          & $0.614 \pm 0.035$          & $0.681 \pm 0.035$          \\ \hline
D-SD        & $0.398 \pm 0.072$         & $0.529 \pm 0.075$         & $0.607 \pm 0.073$         \\ \hline
QD-D        & $0.401 \pm 0.071$         & $0.482 \pm 0.071$         & $0.563 \pm 0.072$         \\ \hline
D-S-DRIL    & $\mathbf{0.576 \pm 0.036}$ & $\mathbf{0.693 \pm 0.033}$ & $\mathbf{0.724 \pm 0.032}$ \\ \hline
D-S-RL      & $0.566 \pm 0.040$          & $0.663 \pm 0.038$          & $0.709 \pm 0.037$          \\ \hhline{|=|=|=|=|}
CTRLSum     & $0.112 \pm 0.046$          & $0.432 \pm 0.073$          & $0.512 \pm 0.076$          \\ \hline
QAGen 2S    & $0.379 \pm 0.071$          & {\ul $0.514 \pm 0.072$}          & {\ul $0.589 \pm 0.072$}          \\ \hline
QA Transfer & {\ul $0.438 \pm 0.140$}          & $0.447 \pm 0.142$          & $0.468 \pm 0.143$          \\ \hline
D-S-NewsQA      & $0.184 \pm 0.108$          & $0.190 \pm 0.119$          & $0.130 \pm 0.097$          \\ \hline
D-S-NQ          & $0.118 \pm 0.108$          & $0.171 \pm 0.125$          & $0.226 \pm 0.147$          \\ \hline
\end{tabular}
}
\caption{Joint accuracy on AT-1, 2 \& 5. \textbf{Bold} represents our best model and \underline{underline} represents best baseline. D-S-DRIL generates significantly better QA pairs than the best performing baseline in all three answer length buckets according to the joint AT-1, 2 \& 5 accuracy.}
\label{tb:hmoverall}
\vspace{-0.2cm}
\end{table}

We report the joint accuracy of \{AT-1, AT-2, AT-5\}, defined by the proportion of QA pairs that are answered \textsc{True} for all three ATs and treat it as a metric for the overall QA pair quality, reporting results in Table \ref{tb:hmoverall} with the following observations. 

\noindent \textbf{D-S-DRIL performs significantly better than the best performing baselines.} The best performing baselines are QA Transfer in length bucket 0 and QAGen 2S in length bucket 1 and 2. We observe that D-D, D-S, D-S-DRIL and D-S-RL all surpass them by a large margin. Particularly, D-S-DRIL outperforms them by $31.51\%$, $34.82\%$ and $22.92\%$ in length bucket $0$, $1$ and $2$, respectively.

\noindent \textbf{DRIL consistently outperforms RL and MLE}. We can see from Table \ref{tb:hmoverall} that D-S-DRIL outperforms D-S and D-S-RL by $3.22\%$ and $2.80\%$, respectively (averaged over $3$ length buckets). The results are consistent on human annotations (AT-2 in Table \ref{tb:qa_matching}, AT-5 in Table \ref{tb:qg_gist}, AT-6 in Figure \ref{fig:ht6}, and joint accuracy in Table \ref{tb:hmoverall}), and automated metrics (QACS in Table \ref{tb:qa_matching} and ROUGE-L/BLEU scores in Table \ref{tb:qg_gist}).
This further shows the advantage of DRIL over MLE and RL, indicating that DRIL can efficiently reinforce AG to generate better QA pairs.

\subsection{Qualitative Analysis}
We also conduct qualitative analyses on generated QA pairs. Please refer to Appendix D for details.

\section{Conclusion}
This paper proposes a model for generating QA pairs with self-contained and summary-centric questions and length-constrained article-summarizing answers. The target application is suggested questions for conversational news recommendation system. We collect a new dataset, \scqa , which contains news articles with questions as titles paired with summaries of varying length. We further propose differential reward imitation learning (DRIL) to efficiently mitigate exposure bias encountered with MLE. Empirically, it is shown that DRIL outperforms multiple alternative baseline neural architectures on automated and human evaluations. 

\section{Broader Impact}
Regarding societal considerations, we consider three aspects. (1) Generating QA pairs that correspond to headlines and article summaries to power a news chatbot can provide users with a rapid glance of recent events. However, exposing users exclusively to article summaries may results in less informed users. Naturally, this can be mitigated by also developing experiences that lead to more in-depth examination of articles, but should be carefully considered. (2) Our \scqa \ dataset collection begins with articles (and potentially news providers) that use questions as article titles. Such articles may have stylistic elements that align with certain forms of journalism (e.g., tabloids) or audience manipulation (e.g., alarmism). Accordingly, the corresponding models may learn to generate similarly biased QA pairs which is certainly undesirable. Future work in this direction may include data cleaning to remove biased QA pairs and/or design de-biased models. (3) Factuality is also a potential issue. A news article itself may be fake news. Meanwhile, the AG model may generate a summary that is factually inconsistent with the corresponding news article. Future work may incorporate recent work in optimizing the factual correctness and considering multiple perspectives of the QA pairs.

\bibliographystyle{acl_natbib}
\bibliography{custom}

\clearpage

\appendix
\section*{Appendix}
In Appendix A, we describe our data collection procedures. In Appendix B, we describe the training details of each algorithm. In Appendix C, we describe the human evaluation setup on Amazon Mechanical Turk. In Appendix D, we provide qualitative analysis of the generated QA pairs of each model.

\section{\scqa: A Self-Contained and Summary-Centric QA Dataset}
In this paper, we propose \scqa , a self-contained and summary-centric QA dataset. The data construction consists of two steps. First, we collect news articles for which their titles are questions, resulting a set of question-article pairs. Second, for each question in the set, we generate $3$ answers that fall into $3$ different length buckets. Details are as follows.

\subsection{Question-Article Collection}
Starting with a curated URL list of news websites, we mined all articles between September 2020 to March 2021 with the following procedure:
\begin{enumerate}
    \item For each news article, we check if the title starts with the following words: `Where', `What', `Did', `Which', `When', `How', `Are', `Is', `Can', `Should', `Who', `Will', `Why', `Whose', `Does', `Do', `Would', `Could', `Shall', `Was', `Were', `Has', `Have', `Had'. If not, filter out that article.
    \item Then we check if the title ends with `?' and not `??'. If not, filter out that article.
    \item If the title matches the following rules, filter out that article: (a) the title includes the word `you', `Stock', etc.\ from an blocklist; (b) the title contain the word `this' which is not followed by a word in a pre-defined allowlist; (c) the title contains stock symbols. We filter out these titles because these are likely clickbait titles. We also filter out titles that contain punctuation marks beside the question mark at the end, as we want the ground-truth questions to be non-complex sentences.
    \item Remove all questions in the articles, as we don't want the model to learn to copy questions from articles.
    \item If the number of tokens in an article is less than $100$, or the number of tokens in the title is less than $3$, filter out that article.
\end{enumerate}

In total, we collected $39{,}461$ question-article pairs.

\subsection{\{Question, Article, Summary, Length Constraint\} 4-Tuples Collection}
Given the collected question-article pairs, we want to augment them with answers of the questions.
We observe that, since the questions are titles of articles, the answers are likely the summaries of articles. From our preliminary study, about $70\%$ of the questions can be answered by the summaries of the corresponding articles. As a result, we propose to augment the question-article pairs with summaries as pseudo ground truth answers. Unfortunately, not all questions can be answered by the generated summaries, this is because (1) even the ground truth summary may not be the correct answer to the question, (2) summaries predicted by the SoTA models are not necessarily good. Therefore, we need a way to identify if a give summary can answer the corresponding question. This is achieved by training a question-answer classifier.

\subsubsection{Question Answer Classifier}
The MS MARCO~\citep{bajaj2016ms} dataset contains $4{,}082{,}910$ labeled question-snippet pairs. A label is either 1 which means that the snippet contains the answer to the question, or 0 which means the snippet does not contain the answer. We fine-tune a classifier based on RoBERTa-large~\citep{liu2019roberta} on the MS MARCO dataset.
To evaluate how good the trained QA classifier is, we generated around $5000$ question-summary pairs, and asked MTurk workers to label whether the summaries answers the corresponding questions. Then we use the trained QA classifier to predict the label.
\begin{table}[h]
\centering
\resizebox{0.9\linewidth}{!}{
\begin{tabular}{|l|l|}
\hline
AUC                  & 0.919                      \\ \hline
Best F1              & 0.960 (P: 0.934, R: 0.989) \\ \hline
F1 at Precision=0.98 & 0.903 (P: 0.980, R: 0.837) \\ \hline
F1 at Precision=0.97 & 0.937 (P: 0.970, R: 0.906) \\ \hline
\end{tabular}
}
\caption{Performance of our QA pair classifier on $5,000$ human annotated QA pairs}
\label{tb:qaclassiferperformance}
\end{table}

The performance of our QA pair classifier is shown in Table \ref{tb:qaclassiferperformance}. We can see that the F1 score of the model is $0.96$ and when the precision is $0.98$, the recall is $0.903$. This shows that the classifier performs sufficiently well for our purposes. Later, we will use this classifier to filter out bad QA pairs. We pick the threshold at which the precision is $0.98$.

\subsubsection{The Length of Answers}
For each question, we want to generate three answers, each contain $1$, $2$ and $3$ sentences. Answers with varying length can accommodate different situations such as different screen sizes of voice assistants.
Table \ref{tb:cnndailymailstats} shows the average number of tokens and characters of the first K sentences in the ground truth summaries of the CNN/DailyMail dataset. \begin{table}[h]
\resizebox{\linewidth}{!}{
\begin{tabular}{|c|c|c|c|c|c|c|c|}
\hline
First K sentence     & 1   & 2   & 3   & 4   & 5   & 6   & 7   \\ \hline
Average \#BPE tokens & 28  & 50  & 72  & 94  & 120 & 140 & 159 \\ \hline
Average \#chars      & 128 & 232 & 336 & 437 & 558 & 655 & 738 \\ \hline
\end{tabular}
}
\caption{Average number of BPE tokens and characters in the first K sentences of ground truth summaries of the CNN/DailyMail dataset.}
\label{tb:cnndailymailstats}
\end{table}
In our work, we define $3$ length buckets for answers with different ranges of number of BPE tokens:
\begin{itemize}
    \item Length bucket 0: (0, 30]
    \item Length bucket 1: (30, 50]
    \item Length bucket 2: (50, 72]
\end{itemize}
Our goal is to be able to specify the length bucket when generate QA pairs, so that we can control the level of brevity for different circumstances (e.g.,\ different screen size of a voice assistant device). 

\subsubsection{Summary (Answer) Generation}
\begin{figure*}[htp!]
\hspace{-1cm}
\centering
\includegraphics[width=0.7\linewidth]{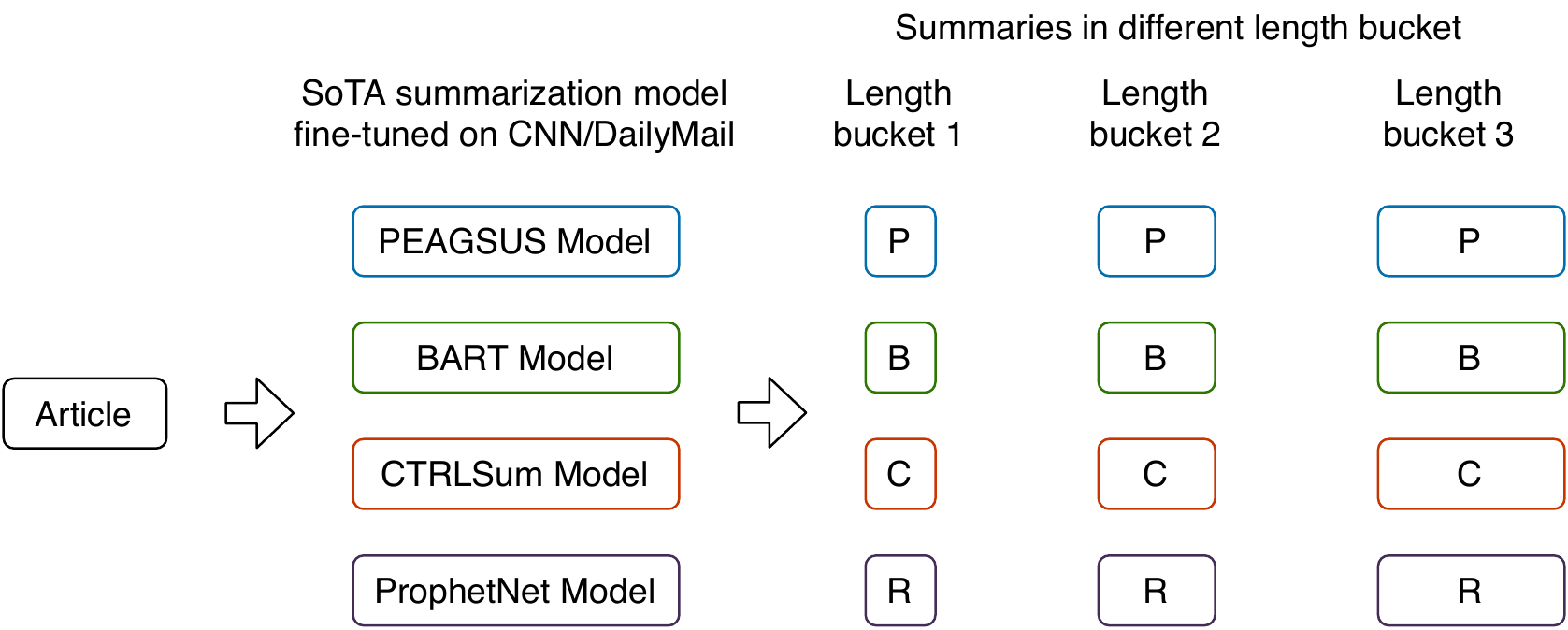}
\caption{SoTA summarization models are used to generate summaries under each length bucket constraints. P, B, C, and R represent the summaries generated by PEAGSUS, BART, CTRLSum and ProphetNet model, respectively.}
\label{fig:summarygen1}
\end{figure*}

\begin{figure*}[htp!]
\includegraphics[width=\linewidth]{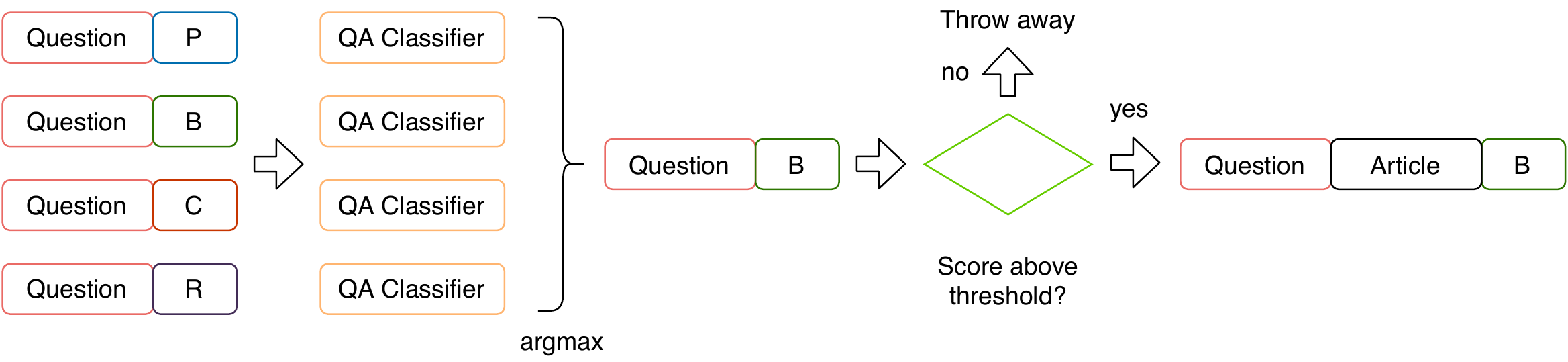}
\caption{Summaries are then scored by the QA pair classifier. The one with the highest score that is also higher than the threshold is kept.}
\label{fig:summarygen2}
\end{figure*}

The high-level idea is to generate summaries using state-of-the-art summarization models under different length constraints and then use the QA pair classifier to filter out unmatched question-summary pairs. The summary generation procedure is shown in Figure \ref{fig:summarygen1}
and Figure \ref{fig:summarygen2}.
In total, we used four summarization models: PEGASUS~\citep{zhang2020pegasus}, BART~\citep{lewis-etal-2020-bart}, CTRLSum~\citep{he2020ctrlsum}, and ProphetNet~\citep{qi-etal-2020-prophetnet}. All models are fine-tuned on CNNDailyMail dataset. However, we found out ProphetNet Model fine-tuned on CNN/DailyMail is uncased\footnote{https://huggingface.co/microsoft/prophetnet-large-uncased-cnndm} so later we removed this model.

For each article, and for each length bucket, we ask each model to generate one summary and we score each question-summary pairs with our QA pair classifier (Note that when generating summaries using CTRLSum, we actually use questions as prompts so that CTRLSum can generate question-conditioned summaries).
To ensure that the generated summaries are in the specified length bucket, we enforce summary length via control of the end-of-the-sentence (\texttt{EOS}) token generation. We remove any unfinished sentences at the end, and then reassign a length bucket.

Finally, for each article and each length bucket, we only keep one summary which has the highest score. We also filter out question-summary pairs which have scores below a threshold (which was chosen so that the QA classifier achieves a precision of $0.98$ as mentioned earlier in this section). 
In Table \ref{tb:summarystats}, we show the number of summaries generated by each model and accepted by our selection strategy. In the future, one could easily introduce more SoTA summarization models in the dataset generation process.
Finally, we generate a dataset containing $53{,}746$ entries. Each entry contains the following components:
question, article, summary, length bucket, QA pair classifier score, model source. Length bucket is an enumerated type consisting of `LB0', `LB1' and `LB2'. Model source is also an enumerated type consisting of `PEGASUS', `BART' and `CTRLSum'. 
Table \ref{tb:summarylenstats} shows the number of BPE tokens and the number of characters of the summaries in each length bucket. Each cell’s format is \#BPE/\#char.

\begin{figure*}[htp!]
\includegraphics[width=\linewidth]{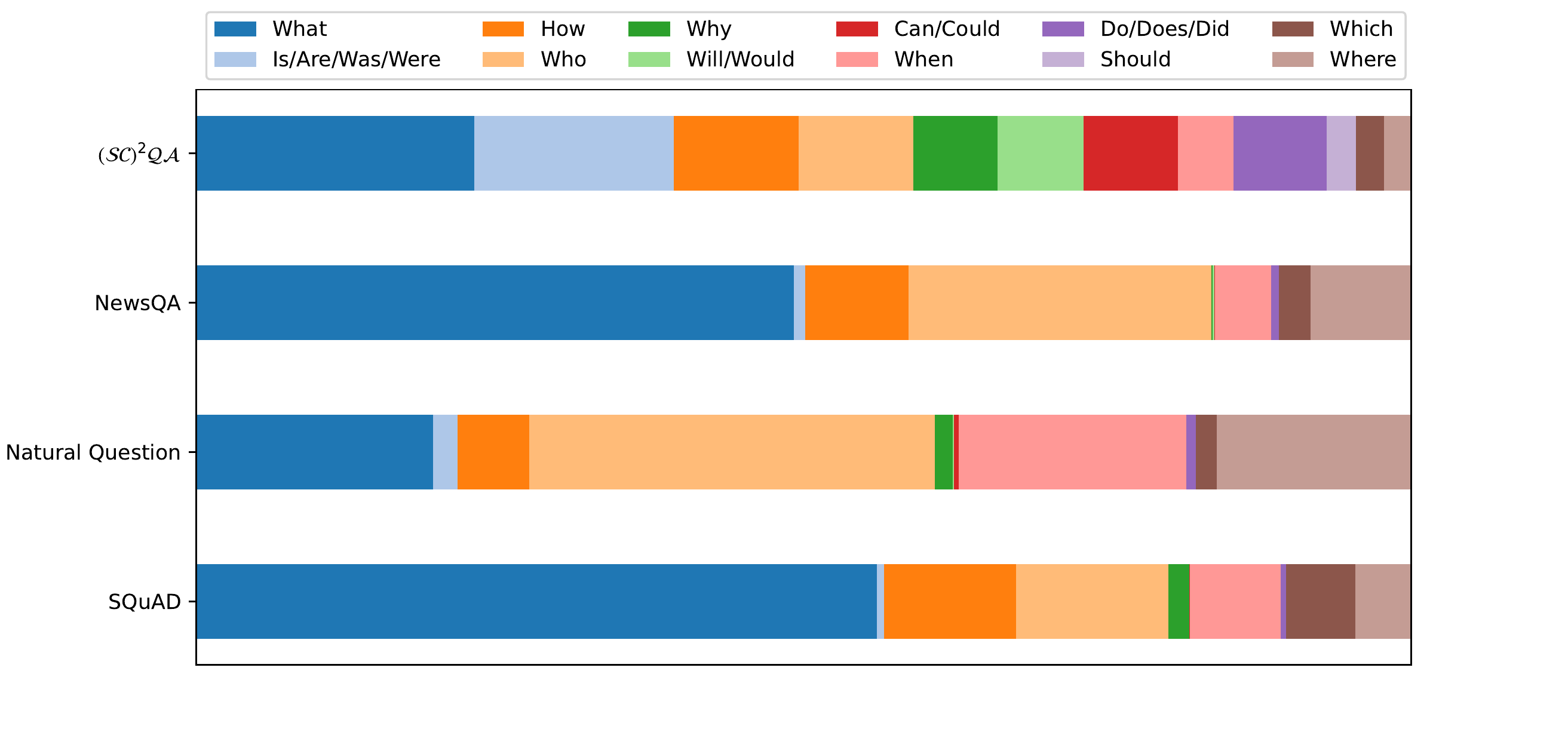}
\caption{Each horizontal bar represents a distribution of the first word of a question in a dataset. Each color represents the proportion of the corresponding word as the first word in a question. The $12$ words shown here are the most frequent first words in all the dataset. This figure shows that \scqa \ has more diverse initial words of questions.}
\label{fig:qdist}
\end{figure*}

Figure \ref{fig:qdist} compares the distributions of the first word of a question in \scqa , NewsQA, Natural Questions, and SQuAD~\citep{rajpurkar-etal-2018-know} dataset. As we can see, \scqa \ is more diverse in terms of the first words in questions.

\begin{table*}[]
\resizebox{\linewidth}{!}{
\begin{tabular}{|l|c|c|c|c|c|c|c|c|c|}
\hline
                                & \multicolumn{3}{c|}{length bucket 0: \#BPE (0, 30{]}} & \multicolumn{3}{c|}{length bucket 1: \#BPE (30, 50{]}} & \multicolumn{3}{c|}{length bucket 2: \#BPE (50, 72{]}} \\ \hline
                                & PEGASUS           & BART           & CTRLSum          & PEGASUS           & BART            & CTRLSum          & PEGASUS           & BART            & CTRLSum          \\ \hline
\#summaries fall in this bucket & 32618             & 31736          & 35638            & 31513             & 30880           & 35030            & 32700             & 34608           & 35719            \\ \hline
\#summaries accepted            & 4112              & 3343           & 4243             & 5542              & 5318            & 7665             & 6807              & 7608            & 9107             \\ \hline
acceptance rate                 & 0.126             & 0.105          & 0.119            & 0.176             & 0.172           & 0.219            & 0.208             & 0.220           & 0.255            \\ \hline
proportion in the dataset       & 0.352             & 0.286          & 0.363            & 0.299             & 0.287           & 0.414            & 0.289             & 0.323           & 0.387            \\ \hline
\end{tabular}
}
\caption{Statistics of summaries generated by each SoTA summarization model in different length buckets.}
\label{tb:summarystats}
\end{table*}

\begin{table*}[]
\resizebox{\linewidth}{!}{
\begin{tabular}{|l|c|c|c|c|c|c|}
\hline
                & min length & max length & mean length  & median length & 10\% percentile length & 90\% percentile length \\ \hline
length bucket 0 & 10/32      & 32/217     & 24.49/105.60 & 25/104        & 18/70                  & 31/143                 \\ \hline
length bucket 1 & 33/88      & 52/339     & 43.56/197.29 & 44/197        & 36/153                 & 51/242                 \\ \hline
length bucket 2 & 53/167     & 74/454     & 63.98/295.18 & 64/293        & 56/224                 & 72/349                 \\ \hline
\end{tabular}
}
\caption{Number of BPE tokens and number of characters of the summaries in each length bucket. Each cell's format is \#BPE/\#char.}
\label{tb:summarylenstats}
\end{table*}

\subsection{Examples of the Data}
Tables \ref{tb:dataexample1} - \ref{tb:dataexample4} show $4$ examples in our dataset.
\begin{table*}[h]
\centering
\resizebox{0.8\linewidth}{!}{
\begin{tabular}{p{\textwidth}}
\hline
\textbf{Article} (truncated): In his first formal White House press conference on Thursday night, President Joe Biden spoke to reporters to outline his plans for immigration, the COVID-19 vaccination effort and foreign policy. He also briefly commented on his own plans for the future, confirming that he does intend to stand for re-election in 2024 and launching some sly digs at his predecessor and the Republican Party. American presidents are limited to two terms in office so almost all choose to stand for a second time. However as the oldest person to be sworn in, there were some doubts as to whether the 78-year-old Biden plans to stand again in 2024. He was directly asked about this at the press conference and answered: ``My plan is to run for re-election, that's my expectation,'' and added that he would fully expect Vice President Kamala Harris to be his running mate again next time around. However he did say that he could not be certain about his plans for the future so soon after taking office, leaving open the possibility that he may decide against a second term. ``Look, I don't know where you guys come from, man,'' he told reporters. ``I'm a great respecter of fate. I've never been able to plan four and a half, three and a half years ahead for certain.'' Biden takes aim at Trump and the GOP Biden has made very few public appearances since taking office in comparison to former President Trump... \\ \hline
\textbf{Question}: What has Biden said about running for re-election in 2024?\\ \hline
\textbf{Summary in length bucket 0}: President Joe Biden made his first formal White House press conference on Thursday night. He confirmed that he plans to stand for re-election in 2024. \\ \hline
\textbf{Summary in length bucket 1}: President Joe Biden made his first formal White House press conference on Thursday night. He confirmed that he plans to stand for re-election in 2024 but left open the possibility that he may decide against a second term.  \\ \hline
\textbf{Summary in length bucket 2}:  President Joe Biden held his first White House press conference on Thursday night. He was asked directly if he plans to run for re-election in 2024. Biden confirmed that he does intend to do so. However he did say that he could not be certain about his plans for the future so soon after taking office. \\ \hline
\end{tabular}
}
\caption{Question-Article-Summary-Length Bucket example 1/5.}
\label{tb:dataexample1}
\end{table*}

\begin{table*}[h]
\centering
\resizebox{0.8\linewidth}{!}{
\begin{tabular}{p{\textwidth}}
\hline
\textbf{Article} (truncated): Public health officials say it's important to vaccinate as many people as quickly as possible to reduce the risk posed by new coronavirus variants. One strategy to stretch existing supplies albeit with huge logistical challenges would be to give just one dose of the vaccine to people who have recovered from COVID-19. About half a dozen small studies, all consistent with one another but as yet unpublished, suggest this strategy could work. Dr. Mohammad Sajadi, at the University of Maryland medical school's Institute of Human Virology studied health care workers who were just getting their first of two vaccine shots. His research team homed in on those who had previously been diagnosed with COVID-19. ``We saw a much faster response and a much higher response,'' he says, based on the protective antibodies his team measured in the blood. The infection served the same priming role as an initial dose of the Moderna or Pfizer vaccine would have, so the first shot they got was in effect a booster. It amplified and solidified immunity to COVID-19. The study was published Monday in JAMA, the journal of the American Medical Association. The Johnson \& Johnson vaccine authorized Saturday by the Food and Drug Administration only requires a single dose. So, he says while vaccine is scarce, it makes sense to offer just one shot to people who have already had the disease. ``You can free up automatically millions of doses,'' he says, increasing vaccine supply by 4 percent or 5 percent. ``We think it makes sense at this time to promote such a policy.'' Federal health officials are intrigued. Dr. Anthony Fauci, who serves as COVID-19 adviser to the White House, has said it's an idea worth further study. He is dead set against another strategy, which is stretching out the time between first and second doses. But health officials are not ready to say yes...  \\ \hline
\textbf{Question}: Could a single-dose of COVID-19 vaccine after illness stretch the supply?\\ \hline
\textbf{Summary in length bucket 0}: One strategy to stretch existing supplies would be to give just one dose of the vaccine to people who have recovered from COVID-19. \\ \hline
\textbf{Summary in length bucket 1}: One strategy to stretch existing supplies would be to give just one dose of the vaccine to people who have recovered from COVID-19. About half a dozen small studies suggest this strategy could work.  \\ \hline
\textbf{Summary in length bucket 2}:  One strategy to stretch existing supplies would be to give just one dose of the vaccine to people who have recovered from COVID-19. About half a dozen small studies suggest this strategy could work. Federal health officials are intrigued, but are not ready to say yes. \\ \hline
\end{tabular}
}
\caption{Question-Article-Summary-Length Bucket example 2/5.}
\label{tb:dataexample2}
\end{table*}

\begin{table*}[h]
\centering
\resizebox{0.8\linewidth}{!}{
\begin{tabular}{p{\textwidth}}
\hline
\textbf{Article} (truncated): Find out in which countries and after what cases vaccination is stopped, what scientists and officials say about the relationship between AstraZeneca and thrombosis, and how the pharmaceutical company itself responded. More than a dozen countries, mostly in the European Union, have suspended the use of the AstraZeneca Covid-19 vaccine due to concerns that some patients have developed blood clots. The World Health Organization (WHO) urged countries to continue using the vaccine, but still decided to convene a meeting due to the massive halt in AstraZeneca vaccination. In total, about 17 million people have received AstraZeneca vaccinations (at least one dose) in the European Union and the UK. Among them, 40 people had blood clots after vaccination. Whether the AstraZeneca vaccine is related to thrombosis is not clear, since its use is not long enough. Vaccine advocates argue that the drug can be used, and the proportion of patients with thrombosis is consistent with the usual statistics, and the vaccine has nothing to do with it. At the same time, many governments have decided to suspend (rather than ban entirely) the vaccination of AstraZeneca pending an investigation by the EMA regulator and estimates by WHO experts. Which countries have suspended vaccination Denmark became the first country to stop using the AstraZeneca Covid-19 vaccine for two weeks after reports of blood clots in some people and even one death on 11 March. A 60-year-old woman who was vaccinated with AstraZeneca developed a blood clot and died. She was vaccinated from the same batch used in Austria. During these two weeks of suspension of vaccinations, the EMA is to investigate. Norway, Iceland, Luxembourg, Romania, and Congo followed Denmark's example. Norwegian authorities said Saturday that four people under 50 who received the AstraZeneca vaccine had unusually low platelet counts in their blood, which could lead to severe bleeding. Bulgaria on March 12 suspended the use of the drug after the death of a 57-year-old woman a few hours after vaccination...  \\ \hline
\textbf{Question}: Why major European nations suspend use of AstraZeneca vaccine?\\ \hline
\textbf{Summary in length bucket 0}: - \\ \hline
\textbf{Summary in length bucket 1}: More than a dozen countries, mostly in the European Union, have suspended the use of the AstraZeneca Covid-19 vaccine due to concerns that some patients have developed blood clots.  \\ \hline
\textbf{Summary in length bucket 2}:  More than a dozen countries, mostly in the European Union, have suspended the use of the AstraZeneca Covid-19 vaccine due to concerns that some patients have developed blood clots. The World Health Organization urged countries to continue using the vaccine, but still decided to convene a meeting due to the massive halt. \\ \hline
\end{tabular}
}
\caption{Question-Article-Summary-Length Bucket example 3/5.}
\label{tb:dataexample3}
\end{table*}

\begin{table*}[h]
\centering
\resizebox{0.8\linewidth}{!}{
\begin{tabular}{p{\textwidth}}
\hline
\textbf{Article} (truncated): Britain's royal family is among the world's most famous organizations -- and a costly one as well. These days, the royal family is known for their lavish weddings, expansive tours and notable fashion as much as they are for their contributions to their nation. According to the BBC, the royals amass their fortune, in part, through the taxpayer-funded Sovereign Grant. However, the queen and the other royals get the money in return for surrendering the profits from their slew of properties -- called the Crown Estate -- to the government, according to Business Insider. Each year, the queen will receive an amount from the grant equivalent to 25\% of the Crown Estate's profits, the outlet reports. The grant will pay for the palace upkeep, the family's travel, royal employee payroll and more, but according to the Telegraph, the Grant doesn't cover costs for security and royal ceremonies, per BI. Money for such assets and events comes from a portfolio of land that the family has owned for generations called the Duchy of Lancaster. The Duchy is made up of residential, commercial, and agricultural properties, Insider reports, and contains \$715 million worth of net assets. In 2019, the portfolio earned \$27 million, The Wall Street Journal reports. The money is put toward `expenses incurred by other members of the royal family,' as the royal family's website puts it...     \\ \hline
\textbf{Question}: Where does the royal family get their money?                  \\ \hline
\textbf{Summary in length bucket 0}: Britain's royal family amass their fortune, in part, through the taxpayer-funded Sovereign Grant. \\ \hline
\textbf{Summary in length bucket 1}: Britain's royal family amass their fortune, in part, through the taxpayer-funded Sovereign Grant. The queen and the other royals get the money in return for surrendering the profits from their slew of properties. \\ \hline
\textbf{Summary in length bucket 2}: Britain's royal family amass their fortune, in part, through the taxpayer-funded Sovereign Grant. The queen and other royals get the money in return for surrendering the profits from their slew of properties to the government. Money for such assets and events comes from a portfolio of land that the family has owned for generations called the Duchy of Lancaster. \\ \hline
\end{tabular}
}
\caption{Question-Article-Summary-Length Bucket example 4/5.}
\label{tb:dataexample4}
\end{table*}

\begin{table*}[h]
\centering
\resizebox{0.8\linewidth}{!}{
\begin{tabular}{p{\textwidth}}
\hline
\textbf{Article} (truncated): NASA's Perseverance rover and its sibling, the Ingenuity helicopter, landed on Mars on February 18, bristling with antennas and cameras. Perseverance, the third robotic visitor from Earth to arrive at the red planet, will spend the next Martian year the equivalent of two Earth years collecting rocks, scrutinizing and photographing them. But the \$2.7-billion robotic explorer has one thing in common with something closer home. The rover has the same processor as the original iMac G3 or the 'Bondi Blue' from 1998. The original iMac used a PowerPC G3 or the PowerPC 750 processor which mirrors the one used in Perseverance, said a report in The Verge. The processor, a single-core, 233MHz processor with just 6 million transistors, was also used in NASA's Curiosity rover, a car-sized rover exploring the red planet which was launched in 2011. The report says that the conditions on Mars could actually be counterproductive for a more advanced processor. Compared to Earth's atmosphere, the atmosphere on the red planet does not offer as much insulation from harmful radiation and charged particles. This could mess up a modern, more complex processor. The Perseverance rover has two computing modules, one being a backup in case of a mishap. Perseverance's processor, a RAD750 chip, is slightly more advanced than the one used in the iMac G3 and is built keeping Mars's radiations in mind. It operates at up to 200 megahertz speed, 10 times the speed in Mars rovers Spirit and Opportunity's computers. Coming to memory power, Perseverance boasts 2 gigabytes of flash memory, 256 megabytes of dynamic random access memory (RAM), and 256 kilobytes of electrically erasable programmable read-only memory. The computer also contains special memory to tolerate the extreme radiation environment that exists in space and on the Martian surface, says NASA...\\ \hline
\textbf{Question}: What do NASA's Mars rover and a 1998 iMac have in common?                  \\ \hline
\textbf{Summary in length bucket 0}: The rover has the same processor as the original iMac G3 or the 'Bondi Blue' from 1998. \\ \hline
\textbf{Summary in length bucket 1}: Perseverance rover has same processor as the original iMac G3 or the 'Bondi Blue' from 1998. The processor was also used in NASA's Curiosity rover, a car-sized rover, launched in 2011. \\ \hline
\textbf{Summary in length bucket 2}: The rover has the same processor as the original iMac G3 or the 'Bondi Blue' from 1998. NASA's Perseverance rover has two computing modules, one being a backup in case of a mishap. The computer also contains special memory to tolerate the extreme radiation environment that exists in space and on the Martian surface, says NASA. \\ \hline
\end{tabular}
}
\caption{Question-Article-Summary-Length Bucket example 5/5.}
\label{tb:dataexample5}
\end{table*}

\section{Training Details}
We use Pytorch and the Transformers package\footnote{https://huggingface.co/transformers/} to implement our algorithms and baselines. The AG models of all the algorithms are initialized by a pre-trained DistilBART model that is fine-tuned on the CNN/DailyMail dataset,\footnote{https://huggingface.co/sshleifer/distilbart-cnn-12-6} and the QG models of all the algorithms are intialized by a pre-trained DistilBART model that is fine-tuned on the XSum dataset.\footnote{https://huggingface.co/sshleifer/distilbart-xsum-12-6} For these two pre-trained models, the number of encoder layers is $12$, the number of decoder layers is $6$, the dimension of hidden states is $1{,}024$, and the number of attention head is $16$.

All the experiments are conducted on AWS EC2 p3dn.24xlarge GPU instances and run with 8 GPUs in parallel. We use the \texttt{Seq2SeqTrainer} from the Transformers package to control the training process. Hyper-parameters are selected based on the ROUGE-L score on validation set described previously (the last $5{,}000$ entries of the data we generated). All the models are optimized with Adam with linear learning rate scheduling, and the number of warm up steps is $500$. All the batch sizes are set to $8$. The number of beams during inference is set to $4$.

\begin{table*}[h]
\centering
\resizebox{\linewidth}{!}{
\begin{tabular}{|l|c|c|c|c|c|c|c|c|}
\hline
            & \multicolumn{4}{c|}{Training}                                                                   & \multicolumn{4}{c|}{Inference}                                                    \\ \cline{2-9} 
            & \multicolumn{2}{c|}{Answer Generation}        & \multicolumn{2}{c|}{Question Generation}        & \multicolumn{2}{c|}{Answer Generation} & \multicolumn{2}{c|}{Question Generation} \\ \cline{2-9} 
            & Encoder                  & Decoder            & Encoder                & Decoder                & Encoder             & Decoder          & Encoder             & Decoder            \\ \hline
D-S         & L + D                    & S                  & S                      & Q                      & L + D               & S'               & S'                  & Q'                 \\ \hline
D-D         & L + D                    & S                  & D                      & Q                      & L + D               & S'               & S'                  & Q'                 \\ \hline
D-SD        & L + D                    & S                  & S + D                  & Q                      & L + D               & S'               & S' + D              & Q'                 \\ \hline
QD-D        & Q + L + D                & S                  & D                      & Q                      & Q' + L + D          & S'               & D                   & Q'                 \\ \hline
D-S-DRIL    & L + D                    & S/S'               & S                      & Q                      & L + D               & S'               & S'                  & Q'                 \\ \hline
D-S-RL      & L + D                    & S/S'                 & S                      & Q                      & L + D               & S'               & S'                  & Q'                 \\ \hhline{|=|=|=|=|=|=|=|=|=|}
QAGen 2S    & D + Q                    & S                  & D                      & Q                      & D + Q'              & S'               & D                   & Q'                 \\ \hline
CTRLSum     & \multicolumn{2}{c|}{Pretrained CTRLSum model} & D                      & Q                      & Q' + D              & S'               & D                   & Q'                 \\ \hline
QA Transfer & Q + D in NewsQA          & A in NewsQA        & D                      & Q                      & Q' + D              & A'               & D                   & Q'                 \\ \hline
D-S-NewsQA  & L + D                    & S                  & D in NewsQA            & Q in NewsQA            & L + D               & S'               & S'                  & Q'                 \\ \hline
D-S-NQ      & L + D                    & S                  & LA in Natural Questions & Q in Natural Questions & L + D               & S'               & S'                  & Q'                 \\ \hline
\end{tabular}
}
\caption{A summary of our models and baselines. Q, S, D, L denote the questions, summaries, documents, and length bucket tags in our dataset, respectively. Q' and S' denote the generated questions and answers, respectively. D in NewsQA, Q in NewsQA, and A in NewsQA denote the documents, questions, and answers in the NewsQA dataset. A' denotes the answers generated by the QA model in QA Transfer. Q + D in NewsQA denotes the concatenation of questions and documents in the NewsQA dataset with \texttt{</s>} as the separator. LA in Natural Questions and Q in Natural Questions denote the long answers and questions in the Natural Questions dataset, respectively.}
\label{table:algosfull}
\end{table*}

\noindent \textbf{D-S}. The QG model's learning rate is $2\times10^{-5}$ and the number of iterations is $5$. The AG model's learning rate is $2\times10^{-5}$ and the number of iterations is $10$.

\noindent \textbf{D-D}. The QG model's learning rate is $3\times10^{-5}$ and the number of iterations is $10$. The AG model is the same as D-S's AG model.

\noindent \textbf{D-SD}. The QG model's learning rate is $3\times10^{-5}$ and the number of iterations is $5$. The AG model is the same as D-S's AG model.

\noindent \textbf{QD-D}. The QG model's learning rate is $3\times10^{-5}$ and the number of iterations is $10$. The AG model's learning rate is $2\times10^{-5}$ and the number of iterations is $10$.

\noindent \textbf{D-S-DRIL}. The QG model is the same as D-S's QG model. The AG model's learning rate is $3\times10^{-5}$ and the number of iterations is $10$. Moreover, as we described in the paper, for the AG model we optimize the sum of DRIL loss and cross entropy loss, and we set $\lambda$ (the weight of the DRIL loss) to $0.3$.

\noindent \textbf{D-S-RL}. The QG model is the same as D-S's QG model. The reward model for AG is a copy of the QG model and is fixed during training. The reward model calculates the negative log-likelihood of a generated question given a generated summary. We use self-critic~\citep{rennie2017self} to train D-S-RL. The learning rate is $2\times10^{-5}$ and the number of iterations is $10$. Similar to D-S-DRIL, we optimize the sum of RL loss and cross entropy loss, and $\lambda$ (the weight of the RL loss) is set to $0.1$.

\noindent \textbf{QAGen 2S}. The learning rate of both the QG and AG model is $2\times10^{-5}$ and the number of iterations is $10$. See Table \ref{table:algosfull} for training and inference pipelines.

\noindent \textbf{CTRLSum}. The QG model is the same as QD-D's QG model. The AG model is the officially pre-trained CTRLSum model.\footnote{https://github.com/salesforce/ctrl-sum} When generating question-conditioned summaries (answers) using the pre-trained CTRLSum model, we use the questions as prompts. See Table \ref{table:algosfull} for training and inference pipelines.

\noindent \textbf{QA Transfer}. The QG model is the same as QD-D's QG model. The AG model is trained on the NewsQA dataset. Since the provided answers in NewsQA dataset are short spans of text, we treat the sentences that contain the answer spans as ground truth answers. The input of the encoder is a concatenation of a question and an article, separated by \texttt{</s>}, and the label of the decoder is the ground truth answer. The learning rate is $2\times10^{-5}$ and the number of iterations is $10$. See Table \ref{table:algosfull} for training and inference pipelines.

\noindent \textbf{D-S-NewsQA}. The QG model is trained on the NewsQA dataset. The input of the encoder is an article, and the label of the decoder is a question. The learning rate is $2\times10^{-5}$ and the number of iterations is $10$. The AG model is the same as D-S's AG model. During inference, questions are generated from summaries. See Table \ref{table:algosfull} for training and inference pipelines.

\noindent \textbf{D-S-NQ}. The QG model is trained on the Natural Questions dataset. The input of the encoder is a long answer, and the label of the decoder is a question. The learning rate is $2\times10^{-5}$ and the number of iterations is $10$. The AG model is the same as D-S's AG model. During inference, questions are generated from summaries. See Table \ref{table:algosfull} for training and inference pipelines.

\section{Human Evaluation Setup}
\begin{figure*}[htp!]
\centering
\includegraphics[width=0.8\linewidth]{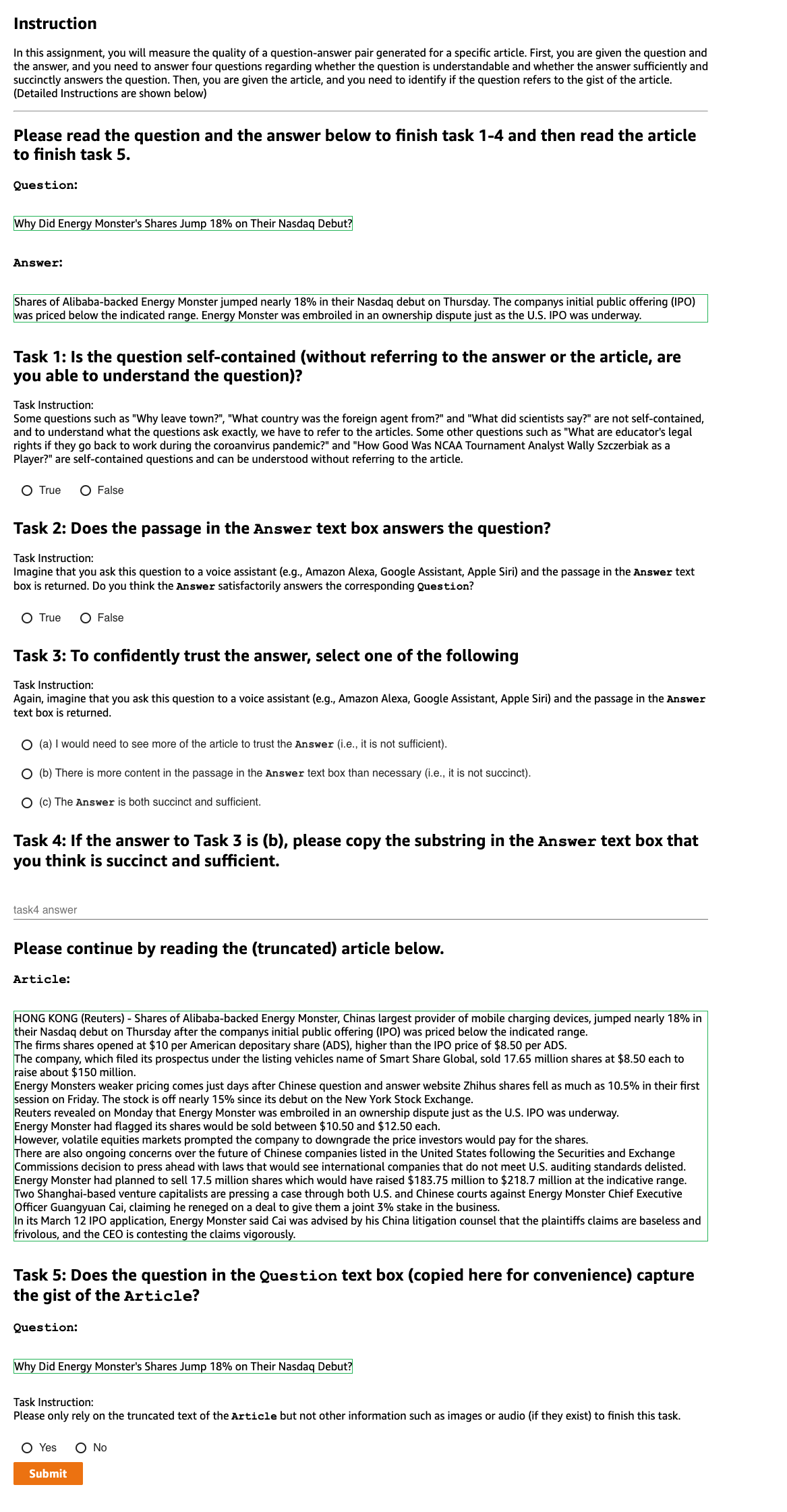}
\caption{Human annotation UI for round 1.}
\label{fig:atfull}
\end{figure*}

We used Amazon Mechanical Turk to conduct human evaluations. In total we completed two rounds of annotation. In round $1$, we evaluated a QA pair generated by a model. The task layout for round $1$ is shown in Figure \ref{fig:atfull}. Each human intelligence task (HIT) has $5$ tasks. First, a QA pair is shown. Task 1 (\textbf{AT-1}) asks if the question is self-contained; Task 2 (\textbf{AT-2}) asks if the answer answers the question; Task 3 (\textbf{AT-3}) asks if the answer is both succinct and sufficient; Task 4 (\textbf{AT-4}) asks the annotator to select a span of the answer that is succinct and sufficient (This task enforces the annotator to read the answers carefully). Following Task 4, we show the corresponding article. Then, Task 5 (\textbf{AT-5}) asks if the question captures the gist of the article.

Each HIT has $3$ assignments, that is, each HIT will be annotated by $3$ different annotators. We used majority vote to aggregate annotations. We designed a qualification task which contains $5$ HITs with their annotations determined by the authors of this paper. We qualified annotators who had an accuracy (using annotations from the authors of this paper as ground truth labels) greater than or equal to $80\%$. We observed that on average it took about $2$ minutes to annotate one HIT. We paid \$$0.35$ per HIT with a \$$0.1$ bonus. We blocked annotators who spent less than $1$ minutes on average on a HIT. If an annotator was blocked, then all the annotations from that annotator were thrown away.

The annotation results in length bucket 0, 1, and 2 are shown in Tables \ref{tb:round1lb0ann} - \ref{tb:round1lb2ann}, respectively. In total, we have $11$ algorithms. During round $1$, we realized that some algorithms performed significantly worse than the others, so there is no reason to collect the equal amount of HITs for every algorithm. Therefore, the number of completed hits for each algorithm is different, as shown in the `completed HITs' columns of Tables \ref{tb:round1lb0ann} - \ref{tb:round1lb2ann}. Meanwhile, since we filtered out annotations from blocked annotators, this also led to different numbers of completed hits between different models. During round 1, we did $7$ mini-round annotations in total (each between $50$ to $150$ HITs), and in the last 3 mini-rounds AT-5 was excluded. When AT-5 was excluded, the annotators did not need to read the article, so the annotation process was accelerated and we were able to collect more annotations for AT-1 to AT-4.

\begin{figure*}[htp!]
\centering
\includegraphics[width=0.6\linewidth]{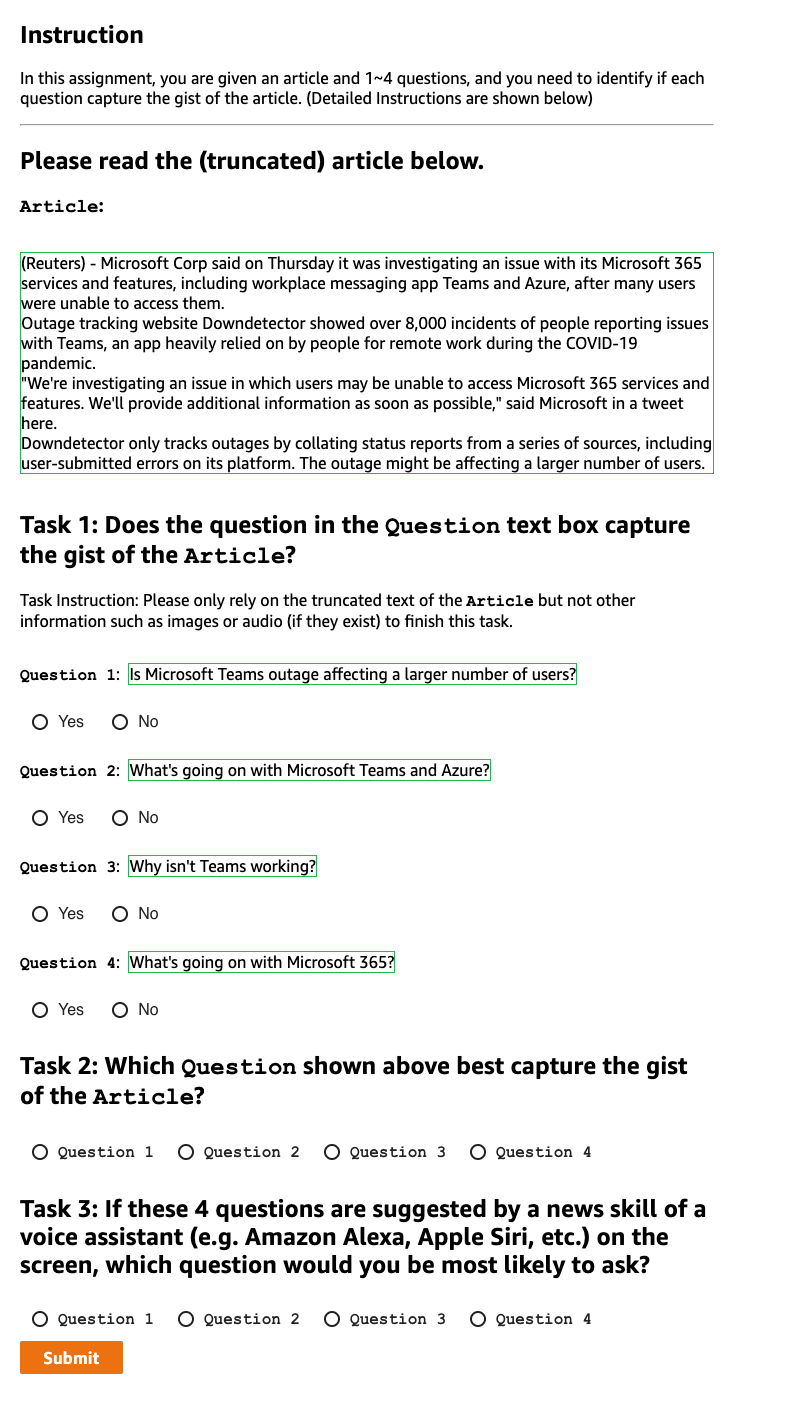}
\caption{Human annotation UI for round 2. Here Task 1 corresponds to AT-5 (same as Task 5 in round 1), Task 2 corresponds to AT-6 and Task 3 corresponds to AT-7.}
\label{fig:attask2}
\end{figure*}

From round 1 we observed that D-S, D-D, D-S-DRIL, and D-S-RL perform the best. Therefore, we conducted annotation round $2$, which compared the questions generated by these four models in one HIT. The task layout for round $2$ is shown in Figure \ref{fig:attask2}. We first show an article, and then show the questions generated by each model. If two or more questions generated by different models are the same, we then merge these questions into one. Therefore, we show $2$ to $4$ questions in one HIT. We randomly shuffle the order of the questions in each HIT, so that the question of a model can appear in any position. Task 1 in round 2 (corresponding to \textbf{AT-5}) asks if each of the question captures the gist of the article; Task 2 in round 2 (corresponding to \textbf{AT-6}) asks which question best capture the gist of the article; Task 3 in round 2 (corresponding to \textbf{AT-7}) asks which question is preferred if suggested by a voice assistant in a news skill. The annotation results in length bucket 0, 1, and 2 are shown in Table \ref{tb:round2lb0ann} - \ref{tb:round2lb2ann}.
While round $1$ and round $2$ both have AT-5, we observe that the three algorithm (D-S, D-D, D-S-DRIL) have lower AT-5 accuracy in round 2 than in round 1. We believe that this is because the round 2 task layout better encourages a more careful reading of the articles by the annotators. However, pairwise preference of AT-5 accuracy is consistent between round 1 and round 2.
\clearpage

\clearpage
\begin{table*}[htp!]
\resizebox{\linewidth}{!}{
\begin{tabular}{|l|c|c|c|c|c|c|c|c|c|c|c|c|c|c|c|}
\hline
            & Completed HITS & AT-1 True & AT-1 False & AT-1 Accuracy & AT-2 True & AT-2 False & AT-2 Accuracy & AT-3 (a) & AT-3 (b) & AT-3 (c) & AT-3 (c) Accuracy & AT-3 (b)+(c) Accuracy & AT-5 True & AT-5 False & AT-5 Accuracy \\ \hline
D-S         & 340            & 311       & 29         & 0.915         & 212       & 128        & 0.624         & 157      & 6        & 167      & {\ul \textbf{0.506}}             & 0.524                 & 144       & 34         & 0.809         \\ \hline
D-D         & 334            & 303       & 31         & 0.907         & 191       & 143        & 0.572         & 163      & 1        & 164      & 0.500               & 0.503                 & 139       & 36         & 0.794         \\ \hline
D-SD        & 176            & 168       & 8          & 0.954         & 71        & 105        & 0.403         & 113      & 1        & 58       & 0.337             & 0.34302               & 157       & 19         & 0.892         \\ \hline
QD-D        & 182            & 168       & 14         & 0.923         & 75        & 107        & 0.412         & 119      & 0        & 58       & 0.328             & 0.328                 & 166       & 16         & {\ul \textbf{0.912}}         \\ \hline
D-S-DRIL    & 377            & 360       & 17         & {\ul \textbf{0.955}}         & 238       & 139        & {\ul \textbf{0.631}}         & 160      & 19       & 174      & 0.493             & {\ul \textbf{0.547}}                 & 144       & 28         & 0.837         \\ \hline
D-S-RL      & 213            & 203       & 10         & 0.953         & 133       & 80         & 0.624         & 99       & 6        & 97       & 0.480             & 0.510                 & -         & -          & -             \\ \hline
CTRLSum     & 178            & 160       & 18         & 0.899         & 20        & 158        & 0.112         & 160      & 2        & 14       & 0.080             & 0.091                 & 153       & 25         & 0.860         \\ \hline
QAGen 2S    & 177            & 158       & 19         & 0.893         & 73        & 104        & 0.412         & 114      & 6        & 50       & 0.294             & 0.329                 & 150       & 27         & 0.847         \\ \hline
QA Transfer & 188            & 177       & 11         & 0.941         & 98        & 90         & 0.521         & 105      & 10       & 63       & 0.354             & 0.410                 & 166       & 22         & 0.883         \\ \hline
D-S-NewsQA  & 241            & 121       & 120        & 0.502         & 148       & 93         & 0.614         & 107      & 22       & 98       & 0.432             & 0.529                 & 12        & 38         & 0.240          \\ \hline
D-S-NQ      & 103            & 86        & 17         & 0.835         & 49        & 54         & 0.476         & 58       & 3        & 38       & 0.384             & 0.414                 & 13        & 37         & 0.260          \\ \hline
\end{tabular}
}
\caption{Round 1 length bucket 0 human annotation. AT-5 annotation for D-S-RL will be in Round 2.}
\label{tb:round1lb0ann}
\end{table*}
\begin{table*}[]
\resizebox{\linewidth}{!}{
\begin{tabular}{|l|c|c|c|c|c|c|c|c|c|c|c|c|c|c|c|}
\hline
            & Completed HITS & AT-1 True & AT-1 False & AT-1 Accuracy & AT-2 True & AT-2 False & AT-2 Accuracy & AT-3 (a) & AT-3 (b) & AT-3 (c) & AT-3 (c) Accuracy & AT-3 (b)+(c) Accuracy & AT-5 True & AT-5 False & AT-5 Accuracy \\ \hline
D-S         & 353            & 335       & 18         & {\ul \textbf{0.949}}         & 264       & 89         & 0.748         & 105      & 87       & 130      & 0.404             & 0.674                 & 161       & 18         & {\ul \textbf{0.899}}         \\ \hline
D-D         & 366            & 341       & 25         & 0.932         & 250       & 116        & 0.683         & 133      & 63       & 128      & 0.395             & 0.590                 & 159       & 27         & 0.855         \\ \hline
D-SD        & 172            & 162       & 10         & 0.942         & 94        & 78         & 0.547         & 81       & 38       & 39       & 0.247             & 0.487                 & 152       & 20         & 0.884         \\ \hline
QD-D        & 193            & 171       & 22         & 0.886         & 98        & 95         & 0.508         & 102      & 36       & 36       & 0.207             & 0.414                 & 165       & 28         & 0.855         \\ \hline
D-S-DRIL    & 380            & 360       & 20         & 0.947         & 293       & 87         & {\ul \textbf{0.771}}         & 104      & 102      & 121      & 0.370             & {\ul \textbf{0.682}}                 & 151       & 17         & {\ul \textbf{0.899}}         \\ \hline
D-S-RL      & 206            & 193       & 13         & 0.937         & 151       & 55         & 0.733         & 64       & 43       & 73       & {\ul \textbf{0.406}}             & 0.644                 & -         & -          & -             \\ \hline
CTRLSum     & 176            & 160       & 16         & 0.909         & 77        & 99         & 0.438         & 104      & 37       & 23       & 0.140             & 0.366                 & 155       & 21         & 0.881         \\ \hline
QAGen 2S    & 183            & 168       & 15         & 0.918         & 98        & 85         & 0.536         & 89       & 37       & 36       & 0.222             & 0.451                 & 166       & 17         & 0.907         \\ \hline
QA Transfer & 189            & 174       & 15         & 0.921         & 111       & 78         & 0.587         & 86       & 35       & 50       & 0.292             & 0.497                 & 162       & 27         & 0.857         \\ \hline
D-S-NewsQA  & 235            & 100       & 135        & 0.426         & 148       & 87         & 0.630         & 93       & 105      & 23       & 0.104             & 0.579                 & 10        & 40         & 0.200         \\ \hline
D-S-NQ      & 110            & 98        & 12         & 0.891         & 56        & 54         & 0.509         & 55       & 21       & 20       & 0.208             & 0.427                 & 18        & 32         & 0.360         \\ \hline
\end{tabular}
}
\caption{Round 1 length bucket 1 human annotation. AT-5 annotation for D-S-RL will be in Round 2.}
\label{tb:round1lb1ann}
\end{table*}
\begin{table*}[]
\resizebox{\linewidth}{!}{
\begin{tabular}{|l|c|c|c|c|c|c|c|c|c|c|c|c|c|c|c|}
\hline
            & Completed HITS & AT-1 True & AT-1 False & AT-1 Accuracy & AT-2 True & AT-2 False & AT-2 Accuracy & AT-3 (a) & AT-3 (b) & AT-3 (c) & AT-3 (c) Accuracy & AT-3 (b)+(c) Accuracy & AT-5 True & AT-5 False & AT-5 Accuracy \\ \hline
D-S         & 337            & 314       & 23         & 0.932         & 254       & 83         & 0.754         & 95       & 125      & 80       & 0.267             & 0.683                 & 153       & 23         & 0.869         \\ \hline
D-D         & 330            & 307       & 23         & 0.930         & 258       & 72         & 0.782         & 86       & 123      & 78       & 0.272             & 0.700                 & 139       & 25         & 0.848         \\ \hline
D-SD        & 173            & 165       & 8          & 0.954         & 113       & 60         & 0.653         & 64       & 65       & 25       & 0.162             & 0.584                 & 151       & 22         & 0.873         \\ \hline
QD-D        & 183            & 163       & 20         & 0.891         & 105       & 78         & 0.574         & 82       & 55       & 27       & 0.165             & 0.500                 & 160       & 23         & 0.874         \\ \hline
D-S-DRIL    & 393            & 381       & 12         & 0.969         & 320       & 73         & {\ul \textbf{0.814}}         & 77       & 181      & 89       & 0.256             & {\ul \textbf{0.778}}                 & 164       & 19         & {\ul \textbf{0.896}}         \\ \hline
D-S-RL      & 208            & 202       & 6          & {\ul \textbf{0.971}}         & 169       & 39         & 0.813         & 41       & 90       & 63       & {\ul \textbf{0.288}}             & 0.777                 & -         & -          & -             \\ \hline
CTRLSum     & 168            & 152       & 16         & 0.905         & 89        & 79         & 0.530         & 83       & 58       & 13       & 0.084             & 0.461                 & 147       & 21         & 0.875         \\ \hline
QAGen 2S    & 180            & 162       & 18         & 0.900         & 110       & 70         & 0.611         & 74       & 61       & 21       & 0.135             & 0.526                 & 156       & 24         & 0.867         \\ \hline
QA Transfer & 195            & 182       & 13         & 0.933         & 134       & 61         & 0.687         & 70       & 72       & 40       & 0.220             & 0.615                 & 165       & 30         & 0.846         \\ \hline
D-S-NewsQA  & 246            & 118       & 128        & 0.480         & 183       & 63         & 0.744         & 60       & 159      & 13       & 0.056             & 0.741                 & 9         & 40         & 0.184         \\ \hline
D-S-NQ      & 100            & 86        & 14         & 0.860         & 67        & 33         & 0.670         & 36       & 42       & 14       & 0.152             & 0.609                 & 16        & 33         & 0.327         \\ \hline
\end{tabular}
}
\caption{Round 1 length bucket 2 human annotation. AT-5 annotation for D-S-RL will be in Round 2.}
\label{tb:round1lb2ann}
\end{table*}

\begin{table*}[]
\centering
\resizebox{0.6\linewidth}{!}{
\begin{tabular}{|l|c|c|c|c|c|c|c|}
\hline
         & AT-5 True & AT-5 False & AT-5 Accuracy & AT-6 Votes & AT-6 Proportion & AT-7 Votes & AT-7 Proportion \\ \hline
D-D      & 255       & 111        & 0.697         & 371        & 0.215           & 386        & 0.224           \\ \hline
D-S      & 281       & 85         & 0.768         & 433        & 0.251           & 424        & 0.246           \\ \hline
D-S-DRIL & 298       & 68         & {\ul \textbf{0.814}}         & 472        & {\ul \textbf{0.273}}           & 457        & {\ul \textbf{0.265}}           \\ \hline
D-S-RL   & 288       & 78         & 0.787         & 450        & 0.261           & 456        & {\ul \textbf{0.265}}           \\ \hline
\end{tabular}
}
\caption{Round 2 length bucket 0 human annotation.}
\label{tb:round2lb0ann}
\end{table*}
\begin{table*}[]
\centering
\resizebox{0.6\linewidth}{!}{
\begin{tabular}{|l|c|c|c|c|c|c|c|}
\hline
         & AT-5 True & AT-5 False & AT-5 Accuracy & AT-6 Votes & AT-6 Proportion & AT-7 Votes & AT-7 Proportion \\ \hline
D-D      & 273       & 98         & 0.736         & 336        & 0.201           & 349        & 0.210           \\ \hline
D-S      & 290       & 81         & 0.782         & 417        & 0.250           & 420        & 0.253           \\ \hline
D-S-DRIL & 299       & 72         & {\ul \textbf{0.806}}         & 466        & {\ul \textbf{0.280}}           & 466        & {\ul \textbf{0.281}}           \\ \hline
D-S-RL   & 289       & 82         & 0.779         & 448        & 0.269           & 426        & 0.256           \\ \hline
\end{tabular}
}
\caption{Round 2 length bucket 1 human annotation.}
\label{tb:round2lb1ann}
\end{table*}
\begin{table*}[]
\centering
\resizebox{0.6\linewidth}{!}{
\begin{tabular}{|l|c|c|c|c|c|c|c|}
\hline
         & AT-5 True & AT-5 False & AT-5 Accuracy & AT-6 Votes & AT-6 Proportion & AT-7 Votes & AT-7 Proportion \\ \hline
D-D      & 287       & 80         & 0.782         & 383        & 0.243           & 383        & 0.246           \\ \hline
D-S      & 300       & 67         & {\ul \textbf{0.817}}         & 413        & {\ul \textbf{0.263}}           & 409        & {\ul \textbf{0.262}}           \\ \hline
D-S-DRIL & 297       & 70         & 0.809         & 401        & 0.255           & 397        & 0.254           \\ \hline
D-S-RL   & 299       & 68         & 0.815         & 376        & 0.239           & 371        & 0.238           \\ \hline
\end{tabular}
}
\caption{Round 2 length bucket 2 human annotation.}
\label{tb:round2lb2ann}
\end{table*}

\clearpage
\section{Qualitative Analysis}
\subsection{Example 1}
\begin{figure*}[htb]
\centering
\fbox{\begin{minipage}[t]{0.97\linewidth}
\textbf{Article} (truncated): \textit{The voices of thousands of college athletes are being heard louder and clearer than they have in years and it is the most politically and socially active generation in a half-centure, since the turbulent years of the late 1960s and early 70s. From seemingly small issues of inequality in NCAA Tournament weight rooms to life-and-death issues of police brutality and endemic racism, athletes are increasingly calling for change, intent on molding what the future should look like for everyone. Some of the things that have occurred this past year, its encouraged a lot of us to speak out on things, social justice, and how we feel, said Loyola Chicagos Lucas Williamson, who is working on a film project involving the schools 1963 national title team that broke down racial barriers. The things weve seen, going back to last summer, its been emotional for me, Williamson said, and its given me the confidence to go out there and speak on some things I feel confident about, and some things that I feel are just causes. While the movement gained momentum last summer, when George Floyd and Breonna Taylor died at the hands of police and protests hit Americas streets, the reality is that social unrest has been bubbling out of sight for years. It took Colin Kaepernick taking a knee to bring it to the surface. The NFL quarterbacks polarizing stance against social and racial injustice in 2016 was embraced by other pro athletes, and that in turn encouraged college athletes to take a stand. They joined the \#MeToo movement against sexual harassment and abuse, and began threatening to strike  to walk off the field of play  unless their demands were heard and met. Protests by more than two dozen Missouri football players against on-campus racism led to he ouster of the president of the university system and the chancellor of its flagship campus. And despite pushback from legislators that threatened to strip funding for scholarships, they found support from athletes on campuses across the country...}
\end{minipage}
} \\
\caption{Article in test set example 1.}
\label{fig:qaae1}
\end{figure*}
The QA pairs generated by each algorithm for the article in Figure \ref{fig:qaae1} are as follows. D-S-DRIL generates the best QA pair in each length bucket. The questions capture the gist of the article and the answers are accurate in all length buckets. On the other hand, D-S, D-S-RL and QAGen 2S' answer in length bucket 1 do not answer the corresponding \texttt{why} question. QA Transfer and CTRLSum generates unfinished sentences (not well-formed) as it cannot control the brevity of the answers.  D-S-NewsQA generates a trivial question in length bucket 0. D-S-NQ's generated questions in length bucket 0 is way off the gist of the article. \\
\textbf{D-S-DRIL Length Bucket 0} [{\ul Question}] Are college athletes' voices getting louder and clearer? [{\ul Answer}] The voices of thousands of college athletes are being heard louder and clearer than they have in years. \\
\textbf{D-S-DRIL Length Bucket 1} [{\ul Question}] Why are college athletes calling for change? [{\ul Answer}] The voices of thousands of college athletes are being heard louder and clearer than they have in years. From seemingly small issues of inequality in NCAA Tournament weight rooms to life-and-death issues of police brutality and endemic racism, athletes are increasingly calling for change. \\
\textbf{D-S-DRIL Length Bucket 2} [{\ul Question}] Why are college athletes calling for change? [{\ul Answer}] The voices of thousands of college athletes are being heard louder and clearer than they have in years. From seemingly small issues of inequality in NCAA Tournament weight rooms to life-and-death issues of police brutality and endemic racism, athletes are increasingly calling for change. They joined the \#MeToo movement against sexual harassment and abuse. \\
\textbf{D-S Length Bucket 0} [{\ul Question}] Are college athletes' voices getting louder and clearer? [{\ul Answer}] The voices of thousands of college athletes are being heard louder and clearer than they have in years. \\
\textbf{D-S Length Bucket 1} [{\ul Question}] Why are college athletes so politically active? [{\ul Answer}] The voices of thousands of college athletes are being heard louder and clearer than they have in years. It is the most politically and socially active generation in a half-centure, since the turbulent years of the late 1960s and early 70s. \\
\textbf{D-S Length Bucket 2} [{\ul Question}] Why are college athletes so politically active? [{\ul Answer}] The voices of thousands of college athletes are being heard louder and clearer than they have in years. It is the most politically and socially active generation in a half-centure, since the turbulent years of the late 1960s and early 70s. From seemingly small issues of inequality in NCAA Tournament weight rooms to life-and-death issues of police brutality and endemic racism, athletes \\
\textbf{D-S-RL Length Bucket 0} [{\ul Question}] Are college athletes ready for the future? [{\ul Answer}] College athletes are increasingly calling for change, intent on molding what the future should look like for everyone. \\
\textbf{D-S-RL Length Bucket 1} [{\ul Question}] Why are college athletes so politically active? [{\ul Answer}] The voices of thousands of college athletes are being heard louder and clearer than they have in years. It is the most politically and socially active generation in a half-centure, since the turbulent years of the late 1960s and early 70s. \\
\textbf{D-S-RL Length Bucket 2} [{\ul Question}] Why are college athletes so politically active? [{\ul Answer}] The voices of thousands of college athletes are being heard louder and clearer than they have in years. It is the most politically and socially active generation in a half-centure, since the turbulent years of the late 1960s and early 70s. From seemingly small issues of inequality in NCAA Tournament weight rooms to life-and-death issues of police brutality and endemic racism, athletes are calling \\
\textbf{D-D Length Bucket 0} [{\ul Question}] What's the latest on college sports news? [{\ul Answer}] The voices of thousands of college athletes are being heard louder and clearer than they have in years. \\
\textbf{D-D Length Bucket 1} [{\ul Question}] Why are college athletes helping to shape politics? [{\ul Answer}] The voices of thousands of college athletes are being heard louder and clearer than they have in years. It is the most politically and socially active generation in a half-centure, since the turbulent years of the late 1960s and early 70s. \\
\textbf{D-D Length Bucket 2} [{\ul Question}] Why are college athletes involved in politics? [{\ul Answer}] The voices of thousands of college athletes are being heard louder and clearer than they have in years. It is the most politically and socially active generation in a half-centure, since the turbulent years of the late 1960s and early 70s. From seemingly small issues of inequality in NCAA Tournament weight rooms to life-and-death issues of police brutality and endemic racism, athletes \\
\textbf{D-SD Length Bucket 0} [{\ul Question}] Why are college athletes speaking out on social issues? [{\ul Answer}] The voices of thousands of college athletes are being heard louder and clearer than they have in years. \\
\textbf{D-SD Length Bucket 1} [{\ul Question}] Why are college athletes speaking out on social issues? [{\ul Answer}] The voices of thousands of college athletes are being heard louder and clearer than they have in years. It is the most politically and socially active generation in a half-centure, since the turbulent years of the late 1960s and early 70s. \\
\textbf{D-SD Length Bucket 2} [{\ul Question}] Why are college athletes speaking out on social issues? [{\ul Answer}] The voices of thousands of college athletes are being heard louder and clearer than they have in years. It is the most politically and socially active generation in a half-centure, since the turbulent years of the late 1960s and early 70s. From seemingly small issues of inequality in NCAA Tournament weight rooms to life-and-death issues of police brutality and endemic racism, athletes \\
\textbf{QD-D Length Bucket 0} [{\ul Question}] Why are athletes speaking out on social issues? [{\ul Answer}] The voices of thousands of college athletes are being heard louder and clearer than they have in years. \\
\textbf{QD-D Length Bucket 1} [{\ul Question}] Why are athletes speaking out on social issues? [{\ul Answer}] The voices of thousands of college athletes are being heard louder and clearer than they have in years. From seemingly small issues of inequality in NCAA Tournament weight rooms to life-and-death issues of police brutality and endemic racism, athletes are calling for change. \\
\textbf{QD-D Length Bucket 0} [{\ul Question}] Why are athletes speaking out on social issues? [{\ul Answer}] The voices of thousands of college athletes are being heard louder and clearer than they have in years. From seemingly small issues of inequality in NCAA Tournament weight rooms to life-and-death issues of police brutality and endemic racism, athletes are calling for change. The movement gained momentum last summer, when George Floyd and Breonna Taylor died at the hands of \\
\textbf{QAGen 2S Length Bucket 0} [{\ul Question}] Why Are College Athletes Calling for Change? [{\ul Answer}] The voices of thousands of college athletes are being heard louder and clearer than they have in years. \\
\textbf{QAGen 2S Length Bucket 1} [{\ul Question}] Why Are College Athletes Calling for Change? [{\ul Answer}] The voices of thousands of college athletes are being heard louder and clearer than they have in years. It is the most politically and socially active generation in a half-centure, since the turbulent years of the late 1960s and early 70s. \\
\textbf{QAGen 2S Length Bucket 2} [{\ul Question}] Why Are College Athletes Calling for Change? [{\ul Answer}] The voices of thousands of college athletes are being heard louder and clearer than they have in years. From seemingly small issues of inequality in NCAA Tournament weight rooms to life-and-death issues of police brutality and endemic racism, athletes are calling for change, intent on molding what the future should look like for everyone. \\
\textbf{CTRLSum Length Bucket 0} [{\ul Question}] Why are athletes speaking out on social issues? [{\ul Answer}] The voices of college athletes are being heard louder and clearer than they have in years. \\
\textbf{CTRLSum Length Bucket 1} [{\ul Question}] Why are athletes speaking out on social issues? [{\ul Answer}] The voices of college athletes are being heard louder and clearer than they have in years. A new generation of athletes is more progressive on such issues as race, gender and \\
\textbf{CTRLSum Length Bucket 2} [{\ul Question}] Why are athletes speaking out on social issues? [{\ul Answer}] The voices of college athletes are being heard louder and clearer than they have in years. A new generation of athletes is more progressive on such issues as race, gender and injustice than previous cohorts at the same age, according to a professor of sociology at Skidmore College \\
\textbf{QA Transfer Length Bucket 0} [{\ul Question}] Why are athletes speaking out on social issues? [{\ul Answer}] According to Andrew M. Linder, a professor of sociology at Skidmore College, there are two main reasons for this athlete-fueled focus on change: First, \\
\textbf{QA Transfer Length Bucket 1} [{\ul Question}] Why are athletes speaking out on social issues? [{\ul Answer}] According to Andrew M. Linder, a professor of sociology at Skidmore College, there are two main reasons for this athlete-fueled focus on change: First, younger people in general are more progressive on such issues as race, gender and injustice than previous cohorts at \\
\textbf{QA Transfer Length Bucket 2} [{\ul Question}] Why are athletes speaking out on social issues? [{\ul Answer}] According to Andrew M. Linder, a professor of sociology at Skidmore College, there are two main reasons for this athlete-fueled focus on change: First, younger people in general are more progressive on such issues as race, gender and injustice than previous cohorts at the same age, and second, they have been emboldened by their athletic heroes. \\
\textbf{D-S-NewsQA Length Bucket 0} [{\ul Question}] What is the name of the college athletes? [{\ul Answer}] The voices of thousands of college athletes are being heard louder and clearer than they have in years. \\
\textbf{D-S-NewsQA Length Bucket 1} [{\ul Question}] What is the most politically active generation in a half centure? [{\ul Answer}] The voices of thousands of college athletes are being heard louder and clearer than they have in years. It is the most politically and socially active generation in a half-centure, since the turbulent years of the late 1960s and early 70s. \\
\textbf{D-S-NewsQA Length Bucket 2} [{\ul Question}] What is the most politically active generation in a half centure? [{\ul Answer}] The voices of thousands of college athletes are being heard louder and clearer than they have in years. It is the most politically and socially active generation in a half-centure, since the turbulent years of the late 1960s and early 70s. From seemingly small issues of inequality in NCAA Tournament weight rooms to life-and-death issues of police brutality and endemic racism, athletes \\
\textbf{D-S-NQ Length Bucket 0} [{\ul Question}] Who is speaking on the college football debate? [{\ul Answer}] The voices of thousands of college athletes are being heard louder and clearer than they have in years. \\
\textbf{D-S-NQ Length Bucket 1} [{\ul Question}] What is the political movement in college football? [{\ul Answer}] The voices of thousands of college athletes are being heard louder and clearer than they have in years. It is the most politically and socially active generation in a half-centure, since the turbulent years of the late 1960s and early 70s. \\
\textbf{D-S-NQ Length Bucket 2} [{\ul Question}] Why are college athletes involved in the civil rights movement? [{\ul Answer}] The voices of thousands of college athletes are being heard louder and clearer than they have in years. It is the most politically and socially active generation in a half-centure, since the turbulent years of the late 1960s and early 70s. From seemingly small issues of inequality in NCAA Tournament weight rooms to life-and-death issues of police brutality and endemic racism, athletes \\

\subsection{Example 2}
\begin{figure*}[htb]
\centering
\fbox{\begin{minipage}[t]{0.97\linewidth}
\textbf{Article} (truncated): \textit{President Biden's infrastructure plan is what this nation has been waiting for, Amtrak chief executive William J. Flynn said, while echoing Biden's push to rebuild and improve the busy Washington-Boston rail corridor. Under the White House plan, intercity rail would receive up to a 400 percent boost in funding, according to some estimates, a transformational investment that could bring major rail expansions and millions more riders. The passenger railroad receives about \$2 billion of federal subsidies annually to cover operations in its national and Northeast networks, as well as other grants and funding for state-sponsored service. The \$2 trillion infrastructure package proposes about \$600 billion of transportation investments, including \$115 billion to rebuild bridges and highways, \$85 billion for transit, \$25 billion to repair and upgrade airports, and \$20 billion for safety initiatives to reduce traffic fatalities. The money, to be spent over eight years, also would address mobility, climate and transportation equity concerns. Amtrak on Wednesday unveiled a plan to provide new intercity rail service to 160 communities and expand service in corridors with heightened demand for rail transportation. The passenger railroad also unveiled a map that highlights 30 possible new routes. The federal funding would help Amtrak along-needed upgrades to tracks, tunnels and bridges in the Northeast, the nations busiest rail corridor. Amtrak has a \$45.2 billion backlog of projects that it says are needed to bring its assets to a state of good repair in the region. Among those projects is the replacement of the Civil War-era Baltimore and Potomac Tunnel in Baltimore, expected to cost \$4.5 billion. Other improvements could be achieved by replacing the North River Tunnels, a more than century-old structure that carries about 200,000 daily passenger trips beneath the Hudson River between New Jersey and New York. An \$11.3 billion plan would double the capacity of existing tunnels, which were damaged by Hurricane Sandy in 2012. Amtrak and other rail services could travel more quickly with the elimination of choke points, additional tracks and other improvements. The passenger railroad has identified about \$18 billion of available or likely to be available funding for projects in the Northeast in the next five years, including the North River Tunnels project...}
\end{minipage}
} \\
\caption{Article in test set example 2.}
\label{fig:qaae2}
\end{figure*}
The QA pairs generated by each algorithm for the article in Figure \ref{fig:qaae2} are as follows. The article is regarding the impact of Biden's infrastructure plan on Amtrak. We can see that the questions generated by D-S-DRIL in length bucket 1 capture the gist of the article, but the questions generated by D-S and D-S-RL in length bucket 1 do not capture the gist of the article. This shows the advantage of the DRIL which generates better summaries. \\
\textbf{D-S-DRIL Length Bucket 0} [{\ul Question}] Is Biden's infrastructure plan the right thing to do for America? [{\ul Answer}] President biden's infrastructure plan is what this nation has been waiting for, Amtrak chief executive William J. Flynn said. \\
\textbf{D-S-DRIL Length Bucket 1} [{\ul Question}] What does President biden's infrastructure plan mean for Amtrak? [{\ul Answer}] President biden's infrastructure plan is what this nation has been waiting for, Amtrak chief executive William J. Flynn said. The federal funding would help Amtrak along-needed upgrades to tracks, tunnels and bridges in the Northeast. \\
\textbf{D-S-DRIL Length Bucket 2} [{\ul Question}] What does President biden's infrastructure plan mean for Amtrak? [{\ul Answer}] President biden's infrastructure plan is what this nation has been waiting for, Amtrak chief executive William J. Flynn said. The federal funding would help Amtrak along-needed upgrades to tracks, tunnels and bridges in the Northeast, the nations busiest rail corridor. Amtrak has a \$45.2 billion backlog of projects that it says are needed to bring its assets to a state of good repair. \\
\textbf{D-S Length Bucket 0} [{\ul Question}] Is biden's infrastructure plan the right thing to do for America? [{\ul Answer}] President biden's infrastructure plan is what this nation has been waiting for, Amtrak chief executive William J. Flynn said. \\
\textbf{D-S Length Bucket 1} [{\ul Question}] What is President biden's infrastructure plan? [{\ul Answer}] President biden's infrastructure plan is what this nation has been waiting for, Amtrak chief executive William J. Flynn said. Intercity rail would receive up to a 400 percent boost in funding, according to some estimates. \\
\textbf{D-S Length Bucket 2} [{\ul Question}] What does President biden's infrastructure plan mean for Amtrak? [{\ul Answer}] President biden's infrastructure plan is what this nation has been waiting for, Amtrak chief executive William J. Flynn said. Intercity rail would receive up to a 400 percent boost in funding, according to some estimates. The federal funding would help Amtrak along-needed upgrades to tracks, tunnels and bridges in the Northeast. \\
\textbf{D-S-RL Length Bucket 0} [{\ul Question}] What is the infrastructure bill? [{\ul Answer}] The \$2 trillion infrastructure package proposes about \$600 billion of transportation investments, including \$115 billion to rebuild bridges and highways. \\
\textbf{D-S-RL Length Bucket 1} [{\ul Question}] What is President biden's infrastructure plan? [{\ul Answer}] President biden's infrastructure plan is what this nation has been waiting for, Amtrak chief executive William J. Flynn said. The \$2 trillion infrastructure package proposes about \$600 billion of transportation investments. \\
\textbf{D-S-RL Length Bucket 2} [{\ul Question}] What is President biden's \$2 trillion infrastructure plan? [{\ul Answer}] President biden's infrastructure plan is what this nation has been waiting for, Amtrak chief executive William J. Flynn said. The \$2 trillion infrastructure package proposes about \$600 billion of transportation investments, including \$115 billion to rebuild bridges and highways, \$85 billion for transit, \$25 billion to repair and upgrade airports. \\
\textbf{D-D Length Bucket 0} [{\ul Question}] Is biden's infrastructure plan the answer to America's infrastructure crisis? [{\ul Answer}] President biden's infrastructure plan is what this nation has been waiting for, Amtrak chief executive William J. Flynn said. \\
\textbf{D-D Length Bucket 1} [{\ul Question}] What would President biden's infrastructure plan mean for Amtrak? [{\ul Answer}] President biden's infrastructure plan is what this nation has been waiting for, Amtrak chief executive William J. Flynn said. Intercity rail would receive up to a 400 percent boost in funding, according to some estimates. \\
\textbf{D-D Length Bucket 2} [{\ul Question}] What would President biden's infrastructure plan mean for Amtrak? [{\ul Answer}] President biden's infrastructure plan is what this nation has been waiting for, Amtrak chief executive William J. Flynn said. Intercity rail would receive up to a 400 percent boost in funding, according to some estimates. The federal funding would help Amtrak along-needed upgrades to tracks, tunnels and bridges in the Northeast. \\
\textbf{D-SD Length Bucket 0} [{\ul Question}] How much federal help is President biden's infrastructure plan for Amtrak? [{\ul Answer}] President biden's infrastructure plan is what this nation has been waiting for, Amtrak chief executive William J. Flynn said. \\
\textbf{D-SD Length Bucket 1} [{\ul Question}] How much would President biden's infrastructure package benefit Amtrak? [{\ul Answer}] President biden's infrastructure plan is what this nation has been waiting for, Amtrak chief executive William J. Flynn said. Intercity rail would receive up to a 400 percent boost in funding, according to some estimates. \\
\textbf{D-SD Length Bucket 2} [{\ul Question}] How much would President biden's infrastructure package benefit Amtrak? [{\ul Answer}] President biden's infrastructure plan is what this nation has been waiting for, Amtrak chief executive William J. Flynn said. Intercity rail would receive up to a 400 percent boost in funding, according to some estimates. The federal funding would help Amtrak along-needed upgrades to tracks, tunnels and bridges in the Northeast. \\
\textbf{QD-D Length Bucket 0} [{\ul Question}] How much would Amtrak get from biden's infrastructure package? [{\ul Answer}] President biden's infrastructure plan is what this nation has been waiting for, Amtrak chief executive William J. Flynn says. \\
\textbf{QD-D Length Bucket 1} [{\ul Question}] How much would Amtrak get from biden's infrastructure package? [{\ul Answer}] President biden's infrastructure plan is what this nation has been waiting for, Amtrak chief executive William J. Flynn said. Under the White House plan, intercity rail would receive up to a 400 percent boost in funding. \\
\textbf{QD-D Length Bucket 2} [{\ul Question}] How much would Amtrak get from biden's infrastructure package? [{\ul Answer}] President biden's infrastructure plan is what this nation has been waiting for, Amtrak chief executive William J. Flynn said. Intercity rail would receive up to a 400 percent boost in funding, according to some estimates. The federal funding would help Amtrak along-needed upgrades to tracks, tunnels and bridges in the Northeast. \\
\textbf{QAGen 2S Length Bucket 0} [{\ul Question}] What does President biden's infrastructure plan mean for Amtrak? [{\ul Answer}] President biden's infrastructure plan is what this nation has been waiting for, Amtrak chief executive William J. Flynn said. \\
\textbf{QAGen 2S Length Bucket 1} [{\ul Question}] What does President biden's infrastructure plan mean for Amtrak? [{\ul Answer}] President biden's infrastructure plan is what this nation has been waiting for, Amtrak chief executive William J. Flynn said. The passenger railroad receives about \$2 billion of federal subsidies annually to cover operations in its national and Northeast networks. \\
\textbf{QAGen 2S Length Bucket 2} [{\ul Question}] What does President biden's infrastructure plan mean for Amtrak? [{\ul Answer}] President biden's infrastructure plan is what this nation has been waiting for, Amtrak chief executive William J. Flynn said. The passenger railroad receives about \$2 billion of federal subsidies annually to cover operations in its national and Northeast networks. The federal funding would help Amtrak along-needed upgrades to tracks, tunnels and bridges in the Northeast. \\
\textbf{CTRLSum Length Bucket 0} [{\ul Question}] How much would Amtrak get from biden's infrastructure package? [{\ul Answer}] President biden's infrastructure plan is what this nation has been waiting for, Amtrak chief executive William J. Flynn says. \\
\textbf{CTRLSum Length Bucket 1} [{\ul Question}] How much would Amtrak get from biden's infrastructure package? [{\ul Answer}] President biden's infrastructure plan is what this nation has been waiting for, Amtrak chief executive William J. Flynn said. The \$2 trillion infrastructure package proposes about \$600 \\
\textbf{CTRLSum Length Bucket 2} [{\ul Question}] How much would Amtrak get from biden's infrastructure package? [{\ul Answer}] The \$2 trillion infrastructure package proposes about \$600 billion of transportation investments, including \$115 billion to rebuild bridges and highways, \$85 billion for transit, \$25 billion to repair and upgrade airports, and \$20 billion for safety initiatives. Amtrak has a \$45 \\
\textbf{QA Transfer Length Bucket 0} [{\ul Question}] How much would Amtrak get from biden's infrastructure package? [{\ul Answer}] The passenger railroad receives about \$2 billion of federal subsidies annually to cover operations in its national and Northeast networks, as well as other grants and funding for state-sponsored service. \\
\textbf{QA Transfer Length Bucket 1} [{\ul Question}] How much would Amtrak get from biden's infrastructure package? [{\ul Answer}] The passenger railroad receives about \$2 billion of federal subsidies annually to cover operations in its national and Northeast networks, as well as other grants and funding for state-sponsored service. \\
\textbf{QA Transfer Length Bucket 2} [{\ul Question}] How much would Amtrak get from biden's infrastructure package? [{\ul Answer}] Under the White House plan, intercity rail would receive up to a 400 percent boost in funding, according to some estimates, a transformational investment that could bring major rail expansions and millions more riders. The passenger railroad receives about \$2 billion of federal subsidies annually to cover operations in its national and Northeast networks, as well as other grants and funding for state-sponsored \\
\textbf{D-S-NewsQA Length Bucket 0} [{\ul Question}] What is the name of the Amtrak chief executive? [{\ul Answer}] President biden's infrastructure plan is what this nation has been waiting for, Amtrak chief executive William J. Flynn said. \\
\textbf{D-S-NewsQA Length Bucket 1} [{\ul Question}] What is the name of the plan? [{\ul Answer}] President biden's infrastructure plan is what this nation has been waiting for, Amtrak chief executive William J. Flynn said. Intercity rail would receive up to a 400 percent boost in funding, according to some estimates. \\
\textbf{D-S-NewsQA Length Bucket 2} [{\ul Question}] What would help Amtrak along? [{\ul Answer}] President biden's infrastructure plan is what this nation has been waiting for, Amtrak chief executive William J. Flynn said. Intercity rail would receive up to a 400 percent boost in funding, according to some estimates. The federal funding would help Amtrak along-needed upgrades to tracks, tunnels and bridges in the Northeast. \\
\textbf{D-S-NQ Length Bucket 0} [{\ul Question}] Who is President's plan for the US infrastructure? [{\ul Answer}] President biden's infrastructure plan is what this nation has been waiting for, Amtrak chief executive William J. Flynn said. \\
\textbf{D-S-NQ Length Bucket 1} [{\ul Question}] Who is the President who wants to build America? [{\ul Answer}] President biden's infrastructure plan is what this nation has been waiting for, Amtrak chief executive William J. Flynn said. Intercity rail would receive up to a 400 percent boost in funding, according to some estimates. \\
\textbf{D-S-NQ Length Bucket 2} [{\ul Question}] How much money would be spent on rail in the White House? [{\ul Answer}] President biden's infrastructure plan is what this nation has been waiting for, Amtrak chief executive William J. Flynn said. Intercity rail would receive up to a 400 percent boost in funding, according to some estimates. The federal funding would help Amtrak along-needed upgrades to tracks, tunnels and bridges in the Northeast. \\

\subsection{Example 3}
\begin{figure*}[htb]
\centering
\fbox{\begin{minipage}[t]{0.97\linewidth}
\textbf{Article} (truncated): \textit{Argentinas President Alberto Fernandez was clear when COVID-19 first hit the country early last year: saving lives at all costs trumped any economic concerns. Now facing a second wave of infections, the South American nation has adjusted its strategy to prioritize protecting its fragile economy. It is hoping greater experience dealing with the coronavirus, a nascent vaccine program, and short, regional lockdowns can help keep the virus in check. The second wave comes at a delicate time for the center-left Peronist government. It is heading for mid-term elections in October to defend its majority in Congress, its popularity bruised by a strict, lengthy lockdown last year and the hard economic hit. The grains producer is also in talks with the International Monetary Fund to revamp some \$45 billion in loans it cannot pay back and needs to fire up economic growth to bring in much needed hard currency. And creditors are looking for signs of recovery after a sovereign debt restructuring last year. The Fernandez administration wants to avoid imposing a blanket lockdown, instead using data on caseloads to establish short-term localized restrictions, reinforce sanitary measures, and maintain controls over borders, a government source said. The government also wants to accelerate a vaccine roll-out delayed by shortage of supply, aiming to have all medical workers and those at high risk vaccinated before the fast-approaching southern winter. Argentinas economy contracted around 10\% last year, the third straight year of recession, and Economy Minister Martin Guzman has said it could not withstand another total shutdown. Poverty levels rose to 42\% in the second half of last year. The country has recorded around 2.4 million coronavirus cases and over 56,000 deaths, and a second wave is building with recent daily cases at 80\% of the peak and rising, a Reuters tally of official data shows. On Tuesday, infections reached a daily record. The second wave and incidence of cases could be even worse when the variants take hold, said Toms Orduna, an infectious disease specialist who advises the government, referring to the P1 Brazilian variant and others racing through the region....}
\end{minipage}
} \\
\caption{Article in test set example 3.}
\label{fig:qaae3}
\end{figure*}
The QA pairs generated by each algorithm for the article in Figure \ref{fig:qaae3} are as follows. The article is about how Argentina dealt with the second wave of the COVID-19 infections. We can see that the QA pairs generated by D-S-DRIL in all length buckets are good\textemdash the questions are self-contained and summary-centric while the answers are correct. Questions generated by QAGen 2S in length bucket 0 and 1 do not have accurate answers. Questions generated by QD-D, D-S-News and D-S-NQ do not capture the gist of the article.   \\
\textbf{D-S-DRIL Length Bucket 0} [{\ul Question}] How is Argentina coping with a second wave of coronavirus? [{\ul Answer}] Argentina facing a second wave of infections, the South American nation has adjusted its strategy to prioritize protecting its fragile economy. \\
\textbf{D-S-DRIL Length Bucket 1} [{\ul Question}] How will Argentina respond to COVID-19 pandemic? [{\ul Answer}] Argentinas President Alberto Fernandez was clear when COVID-19 first hit the country early last year. Now facing a second wave of infections, the South American nation has adjusted its strategy to prioritize protecting its fragile economy. \\
\textbf{D-S-DRIL Length Bucket 2} [{\ul Question}] How is Argentina coping with the second wave of coronavirus? [{\ul Answer}] Argentina has adjusted its strategy to prioritize protecting its fragile economy. It is hoping greater experience dealing with the coronavirus, a nascent vaccine program, and short, regional lockdowns can help keep the virus in check. The second wave comes at a delicate time for the center-left Peronist government. \\
\textbf{D-S Length Bucket 0} [{\ul Question}] Can regional lockdowns keep COVID out of Argentina? [{\ul Answer}] Argentina facing second wave of coronavirus infections. Short, regional lockdowns can help keep virus in check. \\
\textbf{D-S Length Bucket 1} [{\ul Question}] How is Argentina coping with a second wave of coronavirus? [{\ul Answer}] Argentina facing a second wave of infections, the South American nation has adjusted its strategy to prioritize protecting its fragile economy. It is hoping greater experience dealing with the coronavirus and short, regional lockdowns can help keep the virus in check. \\
\textbf{D-S Length Bucket 2} [{\ul Question}] Can regional lockdowns stop the second wave of coronavirus? [{\ul Answer}] Argentina facing second wave of coronavirus infections. Government hopes short, regional lockdowns can help keep virus in check. Second wave comes at delicate time for the center-left Peronist government. It is heading for mid-term elections in October to defend its majority in Congress. \\
\textbf{D-S-RL Length Bucket 0} [{\ul Question}] How is Argentina coping with a second wave of coronavirus? [{\ul Answer}] Argentina facing second wave of coronavirus infections. South American nation has adjusted strategy to prioritize protecting economy. \\
\textbf{D-S-RL Length Bucket 1} [{\ul Question}] How is Argentina coping with a second wave of coronavirus? [{\ul Answer}] Argentina facing a second wave of infections, the South American nation has adjusted its strategy to prioritize protecting its fragile economy. It is hoping greater experience dealing with the coronavirus, a nascent vaccine program, and short, regional lockdowns can help keep the virus in check \\
\textbf{D-S-RL Length Bucket 2} [{\ul Question}] How will the second wave of COVID-19 affect Argentina's economy? [{\ul Answer}] Argentinas President Alberto Fernandez was clear when COVID-19 first hit the country early last year: saving lives at all costs trumped any economic concerns. Now facing a second wave of infections, the South American nation has adjusted its strategy to prioritize protecting its fragile economy. \\
\textbf{D-D Length Bucket 0} [{\ul Question}] How can regional lockdowns help keep COVID-19 out of Argentina? [{\ul Answer}] Argentina facing second wave of coronavirus infections. Short, regional lockdowns can help keep virus in check. \\
\textbf{D-D Length Bucket 1} [{\ul Question}] Can Argentina Keep Coronavirus in Check? [{\ul Answer}] Argentina facing a second wave of infections, the South American nation has adjusted its strategy to prioritize protecting its fragile economy. It is hoping greater experience dealing with the coronavirus and short, regional lockdowns can help keep the virus in check. \\
\textbf{D-D Length Bucket 2} [{\ul Question}] How will regional lockdowns help Argentina keep COVID-19 in check? [{\ul Answer}] Argentina facing second wave of coronavirus infections. Government hopes short, regional lockdowns can help keep virus in check. Second wave comes at delicate time for the center-left Peronist government. It is heading for mid-term elections in October to defend its majority in Congress. \\
\textbf{D-SD Length Bucket 0} [{\ul Question}] How will Argentina deal with COVID-19? [{\ul Answer}] Argentina facing second wave of coronavirus infections. Short, regional lockdowns can help keep virus in check. \\
\textbf{D-SD Length Bucket 1} [{\ul Question}] How will Argentina deal with COVID-19? [{\ul Answer}] Argentina facing a second wave of infections, the South American nation has adjusted its strategy to prioritize protecting its fragile economy. It is hoping greater experience dealing with the coronavirus and short, regional lockdowns can help keep the virus in check. \\
\textbf{D-SD Length Bucket 2} [{\ul Question}] How will Argentina deal with COVID-19? [{\ul Answer}] Argentina facing second wave of coronavirus infections. Government hopes short, regional lockdowns can help keep virus in check. Second wave comes at delicate time for the center-left Peronist government. It is heading for mid-term elections in October to defend its majority in Congress. \\
\textbf{QD-D Length Bucket 0} [{\ul Question}] How will Argentina deal with COVID-19? [{\ul Answer}] Argentina facing second wave of coronavirus infections. Short, regional lockdowns can help keep virus in check. \\
\textbf{QD-D Length Bucket 1} [{\ul Question}] How will Argentina deal with COVID-19? [{\ul Answer}] Argentina facing second wave of coronavirus infections. South American nation has adjusted strategy to prioritize protecting its fragile economy. It is hoping greater experience dealing with the virus, a nascent vaccine program, and short, regional lockdowns can help keep the virus in check \\
\textbf{QD-D Length Bucket 2} [{\ul Question}] How will Argentina deal with COVID-19? [{\ul Answer}] Argentina facing second wave of coronavirus infections. South American nation has adjusted strategy to prioritize protecting economy. Hopes greater experience dealing with virus, a nascent vaccine program, and short, regional lockdowns can help keep virus in check. Second wave comes at delicate time for the center-left Peronist government. \\
\textbf{QAGen 2S Length Bucket 0} [{\ul Question}] How is Argentina dealing with a second wave of coronavirus? [{\ul Answer}] Argentinas President Alberto Fernandez was clear when COVID-19 first hit the country early last year: saving lives at all costs trumped any economic concerns. \\
\textbf{QAGen 2S Length Bucket 1} [{\ul Question}] How is Argentina dealing with a second wave of coronavirus? [{\ul Answer}] Argentinas President Alberto Fernandez was clear when COVID-19 first hit the country early last year: saving lives at all costs trumped any economic concerns. Now facing a second wave of infections \\
\textbf{QAGen 2S Length Bucket 2} [{\ul Question}] How is Argentina dealing with a second wave of coronavirus? [{\ul Answer}] Argentinas President Alberto Fernandez was clear when COVID-19 first hit the country early last year: saving lives at all costs trumped any economic concerns. Now facing a second wave of infections, the South American nation has adjusted its strategy to prioritize protecting its fragile economy. \\
\textbf{CTRLSum Length Bucket 0} [{\ul Question}] How will Argentina deal with COVID-19? [{\ul Answer}] Argentina faces second wave of coronavirus infections. Second wave comes \\
\textbf{CTRLSum Length Bucket 1} [{\ul Question}] How will Argentina deal with COVID-19? [{\ul Answer}] Argentina faces second wave of coronavirus infections. Second wave comes at a delicate time for the center-left Peronist government. Argentina is heading for mid- \\
\textbf{CTRLSum Length Bucket 2} [{\ul Question}] How will Argentina deal with COVID-19? [{\ul Answer}] Argentina faces second wave of coronavirus infections. Second wave comes at a delicate time for the center-left Peronist government. Argentina is heading for mid-term elections in October to defend its majority in Congress, its popularity bruised by a strict, lengthy lockdown \\
\textbf{QA Transfer Length Bucket 0} [{\ul Question}] How will Argentina deal with COVID-19? [{\ul Answer}] The Fernandez administration wants to avoid imposing a blanket lockdown, instead using data on caseloads to establish short-term localized restrictions, reinforce sanitary measures, and maintain controls \\
\textbf{QA Transfer Length Bucket 1} [{\ul Question}] How will Argentina deal with COVID-19? [{\ul Answer}] The Fernandez administration wants to avoid imposing a blanket lockdown, instead using data on caseloads to establish short-term localized restrictions, reinforce sanitary measures, and maintain controls over borders, a government source said. \\
\textbf{QA Transfer Length Bucket 2} [{\ul Question}] How will Argentina deal with COVID-19? [{\ul Answer}] The Fernandez administration wants to avoid imposing a blanket lockdown, instead using data on caseloads to establish short-term localized restrictions, reinforce sanitary measures, and maintain controls over borders, a government source said.  Now facing a second wave of infections, the South American nation has adjusted its strategy to prioritize protecting its fragile economy. \\
\textbf{D-S-NewsQA Length Bucket 0} [{\ul Question}] What is the name of the virus? [{\ul Answer}] Argentina facing second wave of coronavirus infections. Short, regional lockdowns can help keep virus in check. \\
\textbf{D-S-NewsQA Length Bucket 1} [{\ul Question}] What does Argentina want to do to protect its economy? [{\ul Answer}] Argentina facing a second wave of infections, the South American nation has adjusted its strategy to prioritize protecting its fragile economy. It is hoping greater experience dealing with the coronavirus and short, regional lockdowns can help keep the virus in check. \\
\textbf{D-S-NewsQA Length Bucket 2} [{\ul Question}] What is the name of the virus? [{\ul Answer}] Argentina facing second wave of coronavirus infections. Government hopes short, regional lockdowns can help keep virus in check. Second wave comes at delicate time for the center-left Peronist government. It is heading for mid-term elections in October to defend its majority in Congress. \\
\textbf{D-S-NQ Length Bucket 0} [{\ul Question}] What is the cause of the Ebola virus in Argentina? [{\ul Answer}] Argentina facing second wave of coronavirus infections. Short, regional lockdowns can help keep virus in check. \\
\textbf{D-S-NQ Length Bucket 1} [{\ul Question}] What country was hit by Ebola in 2014? [{\ul Answer}] Argentina facing a second wave of infections, the South American nation has adjusted its strategy to prioritize protecting its fragile economy. It is hoping greater experience dealing with the coronavirus and short, regional lockdowns can help keep the virus in check. \\
\textbf{D-S-NQ Length Bucket 2} [{\ul Question}] What is the cause of the virus in Argentina? [{\ul Answer}] Argentina facing second wave of coronavirus infections. Government hopes short, regional lockdowns can help keep virus in check. Second wave comes at delicate time for the center-left Peronist government. It is heading for mid-term elections in October to defend its majority in Congress. \\

\end{document}